\documentclass[11pt]{article}

\usepackage[final]{acl}

\usepackage{times}
\usepackage{latexsym}

\usepackage[T1]{fontenc}

\usepackage[utf8]{inputenc}

\usepackage{microtype}

\usepackage{inconsolata}

\usepackage{graphicx}

\usepackage{amsmath} 
\usepackage{booktabs}
\usepackage{multirow}

\title{Collaborative Multi-Agent Scripts Generation for Enhancing Imperfect-Information Reasoning in Murder Mystery Games}

\author{
Keyang Zhong$^{1,2}$, Junlin Xie$^3$, Hefeng Wu$^1$, Haofeng Li$^1$, Guanbin Li$^{1,2}$\thanks{Corresponding author} 
\\
$^1$Sun Yat-sen University~~~$^2$Shenzhen Loop Area Institute
\\
$^3$The Chinese University of Hong Kong, Shenzhen
\\
zhongky23@mail2.sysu.edu.cn~~~jun0wanan@163.com~~~wuhefeng@gmail.com
\\
lhaof@foxmail.com~~~liguanbin@mail.sysu.edu.cn
}

\begin{document}
\maketitle 
\begin{figure*}[ht]
    \centering
    \includegraphics[width=0.92\textwidth]{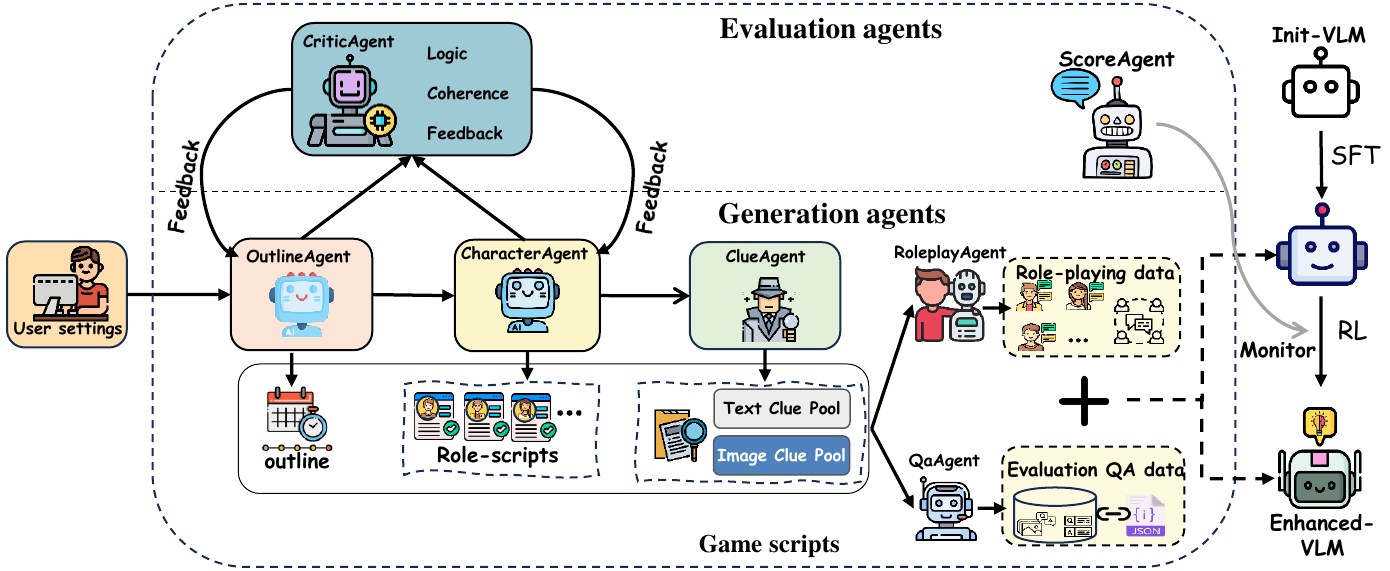}
    \caption{
        Overview of the proposed framework. It employs 
        evaluation agents and generation agents to collaboratively generate logically coherent game scripts and instructs a pretrained VLM via a two-stage training strategy under agent monitoring to enhance the target model's reasoning capability under imperfect information.
    }
    \label{fig:overview}
\end{figure*}

\begin{abstract}
Vision-language models (VLMs) have shown impressive capabilities in perceptual tasks, yet they degrade in complex multi-hop reasoning under multi-player game settings with imperfect and deceptive information. In this paper, we pick up a representative multi-player task, Murder Mystery Games, which require to infer hidden truths based on partial clues provided by the roles of different intentions. To address this challenge, we propose a collaborative multi-agent framework for evaluating and synthesizing high-quality, role-driven multi-player game scripts, enabling fine-grained interaction patterns tailored to character identities (i.e., murderer vs. innocent). Our system generates rich multimodal contexts—including character backstories, visual/textual clues, and multi-hop reasoning chains—through coordinated agent interactions. We design a two-stage agent-monitored training strategy to enhance the reasoning ability of VLM: (1) Chain-of-Thought based fine-tuning on curated and synthetic datasets that model uncertainty and deception; (2) GRPO-based Reinforcement Learning with agent-monitored reward shaping, encouraging the model to develop character-specific reasoning behaviors and effective multi-modal multi-hop inference. Extensive experiments demonstrate that our method significantly boosts the performance of VLM in narrative reasoning, hidden fact extraction, and deception-resilient understanding. Our contributions offer a scalable solution for training and evaluating VLMs under uncertain, adversarial, and socially complex conditions, laying the groundwork for future benchmarks in multimodal multi-hop reasoning under imperfect information. 
\end{abstract}
\section{Introduction}

Vision-language models (VLMs) have demonstrated impressive capabilities in foundational perceptual tasks such as image captioning and visual question answering (VQA), as well as in more complex reasoning tasks through chain-of-thought (CoT) prompting, leveraging their ability to align and integrate information across visual and linguistic modalities \cite{openai_gpt4v_system_card_2023, google_gemini_ai_2023, li2025survey,wang2025multimodalchainofthoughtreasoningcomprehensive}. However, tasks that demand sophisticated reasoning particularly those involving multi-hop inference, imperfect or deceptive information, and dynamic social interactions—remain challenging \cite{yang2018hotpotqa,chen2024socialbenchsocialityevaluationroleplaying,chen2024deceptiondetectiondeeperdataset}. Recent efforts have begun to explore more expressive and controllable reasoning paradigms, to overcome bottlenecks in representation and reasoning capacity \cite{dong2025aurora,dong2026neureasonerexplainablecontrollableunified,jiang2026foeforesterrorsmakes}.
Advancing VLMs in such settings necessitates evaluation and training environments that require not only perception and knowledge, but also deeper reasoning and adaptability under imperfect information.

In real life, many practical tasks involve a multi-player game-theoretic process using imperfect information. For example, in judicial proceedings, judges, prosecutors, defense lawyers, witnesses, and juries engage in multiple rounds of social interaction from their respective perspectives, and make multi-hop inferences and decisions based on incomplete information, ultimately attempting to finish their own task.

To study such imperfect-information multi-player reasoning in vision-language models, we adopt Murder Mystery as a representative test environment. Murder Mystery is a social deduction role-playing game in which players assume predefined identities and collaboratively infer the hidden murderer, making it a typical yet tractable setting for modeling multi-agent interaction and reasoning under uncertainty.  With access to public and private textual and visual clues, players engage in structured dialogue to reason about motives and inconsistencies, and infer the murderer amid adversarial deception. The game proceeds through four key phases:

\begin{enumerate}
    \item \textbf{Role Setup and Clue Absorption:} Players receive the rules and character backgrounds, followed by textual and visual clues. They then provide in-character self-introductions as the begin.
    \item \textbf{Interactive Discussion:} Players engage in question-and-answer interaction with each other based on their clues and suspicions, emphasizing social inference, inconsistency detection, credibility assessment, and information selection.
    \item \textbf{Hypothesis Generation:} Integrating accumulated clues and dialogue, players generate reasoning chains to infer motives and methods. This phase requires multi-hop multimodal reasoning across narrative and visual content.
    \item \textbf{Final Decision:} Each player makes a final judgment regarding the murderer's identity.
\end{enumerate}

This setting embodies key challenges such as imperfect information, inconsistency detection, and strategic social interaction, making it a suitable testbed for evaluating multi-hop multimodal reasoning in vision-language models. In addition, the task probes models’ abilities in long-form narrative understanding, multimodal evidence integration, and the synthesis of textual and visual information through multi-step inference.

Although Murder Mystery is a representative task for modeling multi-player game process, there is still a lack of large-scale datasets for fine-tuning and evaluating models. Large-scale production of high-quality murder mystery scripts is expensive and impractical. 
To address this challenge, we design a multi-agent simulation framework where powerful LLMs (e.g., Gemini 2.5Pro \cite{gemini2025report}) act as autonomous agents to collaboratively synthesize diverse Murder Mystery game scripts, producing challenging questions-answering pairs and multi-player interactive dialogue as training datasets\cite{ma2026talk2image,zhang-etal-2025-ga}. 
To enable the fine-tuning of VLMs on complex and adversarial examples, we build a new paradigm to automatically generate reasoning chains based on incomplete information. Lastly, we adopt a scalable two-stage training pipeline to learn VLMs\cite{han2026unicorn}, via combining high-quality, auto-generated cases with curated training data.

Our main contributions are summarized as follows:
\begin{itemize}
    \item \textbf{Multi-Agent Script Synthesis Framework:} We propose a scalable multi-agent framework to automatically generate diverse, high-quality multi-player game scripts. This framework simulates realistic character roles, player interactions and multimodal clues. 
    \item \textbf{Training Data Construction and Learning under Imperfect Information:} We develop a novel paradigm for generating reasoning chains under imperfect information, enhancing model learning via a two-stage agent-monitored strategy. 
    \item \textbf{Performance Enhancement under Imperfect Information:} Our method demonstrates consistent performance gains in reasoning and role-playing for vision-language models at both the 3B and 7B scales (e.g., Qwen2.5-VL-3B-Instruction and its 7B counterpart) in Murder Mystery scenarios involving imperfect and deceptive information.

\end{itemize}

\section{Related Work}
\paragraph{Social Reasoning Games as VLM Evaluation Platforms}

Social reasoning games, such as Murder Mystery and Werewolf, have become prominent platforms for evaluating the reasoning capabilities of VLMs in settings characterized by imperfect information, multi-agent interactions, and deception. These games provide a robust framework for assessing cognitive resilience in complex, multimodal scenarios \cite{zhu2025playerenhancingllmbasedmultiagent,wu2024enhancereasoninglargelanguage}.
WhodunitBench, offers 50 murder myster scripts with both multiple-choice and open-ended questions to facilitate multi-agent reasoning assessment \cite{xie2024whodunitbench}. Frameworks such as MultiMind extend the evaluation to non-verbal modalities, incorporating facial expressions and intonation \cite{zhang2025multimind}.
The SocialMaze benchmark focuses on VLM reasoning in static social contexts, explicitly excluding deceptive elements \cite{xu2025socialmazebenchmarkevaluatingsocial}. Other frameworks, including BALROG, KORGym, and VS-Bench, assess multimodal reasoning in dynamic game environments but do not explicitly target social interaction capabilities \cite{paglieri2024balrog,shi2025korgymdynamicgameplatform,xu2025vsbenchevaluatingvlmsstrategic}. 

\paragraph{Multi-Agent Synthetic Data}

The scarcity of high-quality multimodal training data remains a significant bottleneck for VLM development. 
Synthetic data generation, particularly through multi-agent systems, has emerged as a scalable solution that enhances dataset diversity and reasoning complexity while reducing reliance on manual annotation\cite{ma2026talk2image,zhang-etal-2025-ga}.
AgentInstruct utilizes a hierarchical multi-agent workflow to automatically produce synthetic instruction-response data with minimal human involvement \cite{mitra2024agentinstructgenerativeteachingagentic}. 
Similarly, MATRIX simulates multi-agent social scenarios to generate data for alignment and instruction tuning \cite{tang-etal-2025-synthesizing}.
AudioGenie uses a dual-team multi-agent framework consisting of a "generation team" and a "supervision team" to generate diverse audio from multimodal inputs\cite{rong2025audiogenie}.
Frameworks such as GenArtist and LayerCraft operate on similar principles. GenArtist decomposes complex text prompts into sub-tasks using a VLM-based agent, constructs detailed planning trees, and leverages external tools (e.g., SDXL, DALL-E 3) for image generation and editing. Iterative verification and self-correction further enhance output fidelity \cite{wang2024genartist,zhang2025unified,zhang2025layercraft}. 
Recent works on composed image retrieval further highlight the importance of modeling fine-grained modification signals and compositional semantics for generating high-quality multimodal data \cite{HABIT,OFFSET,INTENT,HINT,MELT}.

\paragraph{Training Pipelines for Reasoning-Enhanced VLMs}

Recent research frequently adopts a supervised fine-tuning (SFT) followed by reinforcement learning (RL) pipeline to enhance VLM reasoning. Both Reason-RFT \cite{tan2025reasonrftreinforcementfinetuningvisual} and SRPO \cite{zhang2025srpocrossdomainimplementationlargescale} employ this two-stage approach: SFT is used to instill structured chain-of-thought reasoning, while RL further optimizes reasoning quality and generalization\cite{dong2025aurora,dong2026neureasonerexplainablecontrollableunified,jiang2026foeforesterrorsmakes}.
In a curriculum-based paradigm, Infi-MMR \cite{liu2025infimmrcurriculumbasedunlockingmultimodal} progressively transitions from textual to multimodal and caption-free reasoning using sequential RL, achieving strong results on multimodal math benchmarks. VILASR introduces a “drawing-to-reason” paradigm, utilizing simple visual operations (e.g., auxiliary lines) to articulate spatial relationships, and employs a three-stage training process—synthetic data pre-training, reflective rejection sampling, and RL—to improve self-correction and generalization \cite{wu2025reinforcingspatialreasoningvisionlanguage}.
Recent efforts also explore retrieval-augmented and experience-driven learning paradigms to improve long-horizon reasoning and interaction efficiency. For instance, ExpSeek proposes a self-triggered experience seeking mechanism for web agents, enabling adaptive data acquisition and policy refinement during training \cite{zhang2026expseek}. Meanwhile, self-supervised and self-improving training paradigms further push toward unified multimodal learning without heavy human annotation \cite{han2026unicorn}.

\begin{figure*}[!t]
    \centering
    \includegraphics[width=1.0\textwidth]{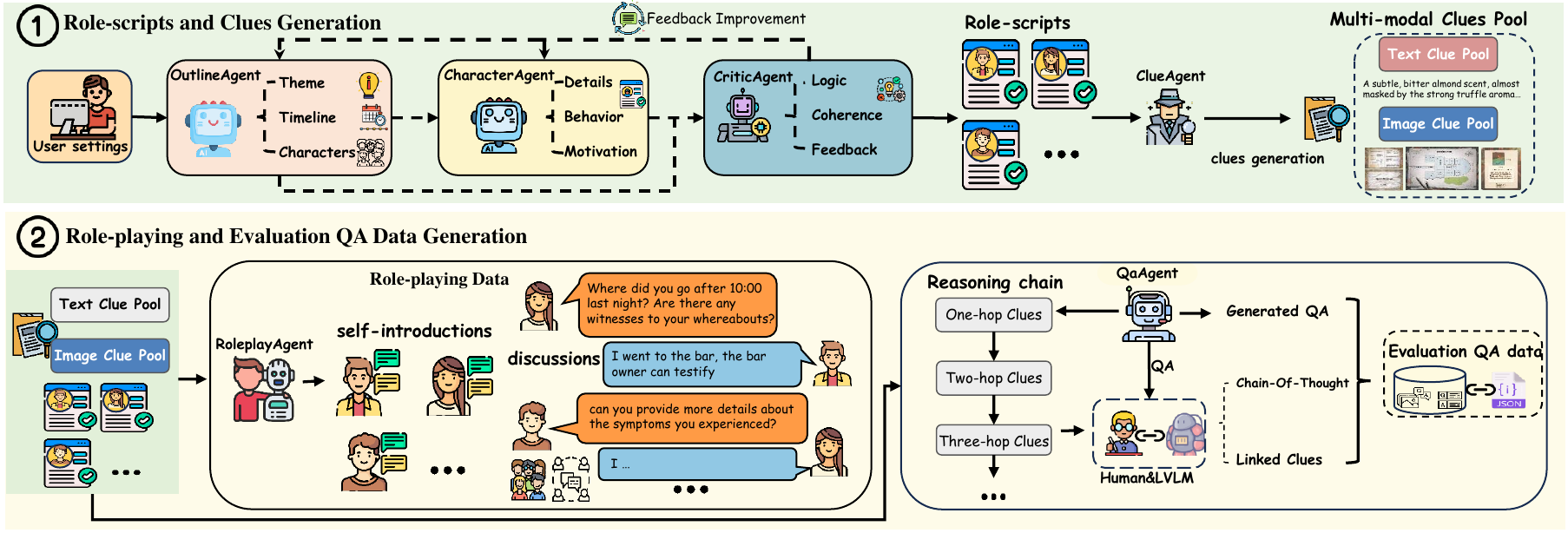}
    \caption{
        The details of game scripts generated via our multi-agent framework.
    }
    \label{fig:dataGeneration} 
\end{figure*}

\section{Method} 

This section first describes our multi-agent framework, and then takes the  Murder Mystery Games as the application scenario to depict the process of applying our framework.

\subsection{Overview of Proposed Multi-Agent Framework}

The proposed collaborative multi-agent framework aims to leverage collaborative agents to generate high-quality training data and instruct a pretrained VLM to enhance its reasoning under imperfect information in game-theoretic tasks.Our multi-agent framework includes two types of agents, i.e., \textbf{generation agents} and \textbf{evaluation agents}. While generation agents simulate realistic, interactive game processes to generate game-script data, evaluation agents focus on assessing the quality of these generated outputs and offering constructive feedback for improvement.

As shown in Figure \ref{fig:overview}, the framework includes generation agents such as the OutlineAgent, which produces story outlines with background and role summaries; the CharacterAgent, which creates detailed role scripts; and the ClueAgent, which generates multimodal clues that convey key environmental information. Building upon these elements, the RoleplayAgent produces role-playing data for specific scenarios, while the QaAgent constructs question–answer pairs to assess and strengthen the model’s reasoning ability. To ensure quality, the CriticAgent evaluates the generated scripts for logical coherence and behavioral consistency. During training, the ScoreAgent assesses the model’s role-specific behaviors, measuring how well its interactions align with the designated roles, and uses this feedback to facilitate model improvement. All agents interact through shared game scripts, working collaboratively to enhance imperfect-information reasoning, and the framework remains extensible for diverse game-theoretic tasks by adapting or adding specialized agents. 

\paragraph{Application to Murder Mystery Games}
Although game-theoretic tasks are widely present in social life, there are rare well-defined benchmarks for enhancing VLMs' reasoning ability in such scenarios. Recently, a benchmark \cite{xie2024whodunitbench} rooted from Murder Mystery Games has emerged as a VLM evaluation platform. Though its data remains insufficient, it offers a well-defined game-theoretic scenario.
Therefore, we validate the soundness of the proposed framework by training models on data synthesized by our framework and evaluating their effectiveness in Whodunitbench.

Under the Murder Mystery Games evaluation protocol, a VLM receives a context defined as $ \mathcal{C} = \{\mathbf{B}, \mathbf{I}, \mathbf{T}, \mathbf{D}\} $, where $\mathbf{B}$ denotes character backgrounds, $\mathbf{I} = \{I  _n\}_{n=1}^N$ is a set of image-based clues, $\mathbf{T} = \{T_m\}_{m=1}^M$ comprises public textual clues, and $\mathbf{D}$ captures the dialogue history. The model is required to demonstrate role-playing fidelity, detect deception by other participants, and execute sophisticated reasoning over multimodal clues.

\begin{figure*}[ht]
    \centering
    \includegraphics[width=1.0\linewidth]{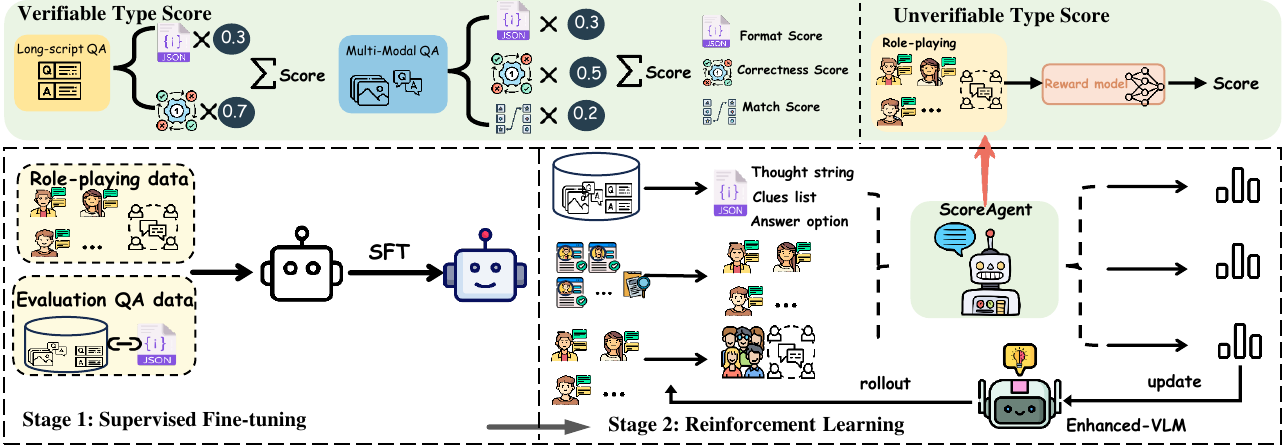}
    \caption{The bottom part outlines the two-stage training strategy. The top part showcases the ScoreAgent design, which applies specialized reward functions to different training data types for reward calculation during reinforcement learning. }
    \label{fig3:trainingframework}
\end{figure*}

\subsection{Agent-Driven High-Quality Data Generation}
\label{sec:3.1scripts-data-synthesis}
Each agent is instantiated via carefully designed prompts to a strong proprietary model(Figure \ref{fig:dataGeneration}).
OutlineAgent first constructs the crime-day narrative with basic motives and secrets. CharacterAgent elaborates detailed daily actions and interactions while maintaining suspense.
CriticAgent evaluates the resulting scripts across four dimensions: plot complexity, character development, difficulty, and logical rationality, and gives feedback for refinement.
ClueAgent produces multimodal clues—visual or textual—that aid deduction without revealing the culprit.
RoleplayAgent then simulates multi-turn dialogues, while QaAgent generates reasoning chains and QA pairs (from one-hop to multi-hop) with annotated step-by-step reasoning and supporting evidence.

The resulting training corpus consists of two components: interactive role-playing data, providing context-rich dialogue trajectories, and structured QA data covering perception and cognition, with explicit reasoning and evidence grounding. Together, these data support subsequent agent-monitored model enhancement, enabling robust and multifaceted capability injection into the target VLM. Detailed agent specifications and dataset descriptions are provided in Appendices A and B.

\subsection{Agent-Monitored Model Enhancement}
\label{sec:3.2TrainingFramework}

To effectively enhance the target model's reasoning capability under imperfect information, we adopt a two-stage training strategy: (i) direct fine-tuning with synthetic offline data to establish basic role-playing and reasoning capabilities in Murder Mystery Games, and (ii) GRPO-based reinforcement learning monitored by ScoreAgent to incentivize reasoning potentials, as illustrated in Figure \ref{fig3:trainingframework}.

\subsubsection{Supervised Fine-tuning}
Since the training data are synthesized by agents built on powerful large-scale VLMs, fine-tuning allows a smaller target model to inherit structured reasoning patterns and role-playing interaction behaviors, leading to improved performance on complex multimodal inference tasks.
In practice, we apply parameter-efficient fine-tuning with LoRA\cite{hu2021loralowrankadaptationlarge} to the pretrained VLM using standard autoregressive supervision over the generated answers and reasoning traces.

\subsubsection{Reinforcement Learning} 
\begin{figure}[!t]
    \centering
    \includegraphics[width=1.0\linewidth]{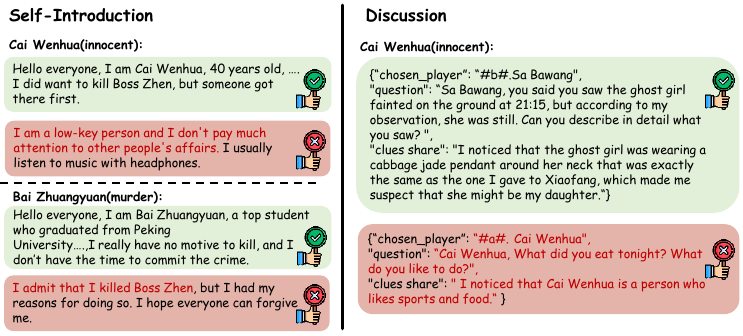}
    \caption{Red denotes low-scoring and green denotes high-scoring examples. Low scores arise from off-topic, self-contradictory, or rule-violating behaviors in self-introductions and discussions, such as irrelevance to the script or premature identity revelation.}
    \label{fig4:accept_reject}
    \vspace{-1ex}
\end{figure}
Murder Mystery Games involve role-consistency behavior under imperfect information, requiring truthful cooperation from innocent players and strategic deception from the murderer—a setting where existing VLMs exhibit clear limitations. Since self-introductions and discussions do not have a single correct answer and are highly context-dependent, supervised fine-tuning (SFT) alone is insufficient to meet these requirements.
Instead of training an independent reward model, we adopt an LLM-as-Judge approach \cite{zhu2023judgelm,li2025generationjudgmentopportunitieschallenges,whitehouse2025j1incentivizingthinkingllmasajudge}. Specifically, the ScoreAgent introduced earlier is implemented by prompting a powerful LLM to score candidate interactions—assigning higher rewards to identity-consistent self-introductions and meaningful discussions, while assigning lower rewards to irrelevant or rule-violating responses, as illustrated in Figure \ref{fig4:accept_reject}.

For evaluation on unverifiable data types , i.e., generated self-introduction and discussion where no ground truth exists, we design the reward as:
\begin{equation}
S = 
\begin{cases}
\text{M}(\mathcal{R
}), & \\ \text{for Self-Introduction}, \\
\gamma \text{M}(\mathcal{R
}) + (1-\gamma) S_{\text{choice}}(\mathcal{R
}), & \\ \text{for Discussion}.
\end{cases} 
\end{equation}
We define the auxiliary reward components as follows: $S_{\text{format}}(\mathcal{R
})$ equals 1 if the response $\mathcal{R
}$ is in valid JSON format, and 0 otherwise; $S_{\text{match}}(\mathcal{R
})$ equals 1 if all referenced image clues are correctly matched, and 0 otherwise; $S_{\text{choice}}(\mathcal{R
})$ equals 1 if the player chooses to ask the murder suspect, 0.5 if asking another player, and 0 if asking themselves.

For evaluation on verifiable data type (e.g., with standard answers for long script QA and multimodal QA), we define the reward functions as follows:
\begin{equation}
S = 
\begin{cases}
\mathbf{1}(\alpha S_{\text{correct}} > \beta) \cdot \big[ S_{\text{correct}}(A,\hat{A}) \\
\qquad + (1-\alpha) S_{\text{format}}(\mathcal{R
}) \big], \text{(a)}, \\[10pt]
\mathbf{1}(\alpha S_{\text{correct}} > \beta) \cdot \bigg[ S_{\text{correct}}(A,\hat{A}) \\
 + \frac{1}{2}(1-\alpha)\big( S_{\text{format}}(\mathcal{R
}) + S_{\text{match}}(\mathcal{R
}) \big) \bigg],  \text{(b)}.
\end{cases}
\end{equation}
(a) for Long Script QA and (b) for Multimodal QA. Here, $\alpha$ balances answer correctness with format and clue matching, while $\beta$ imposes a correctness threshold to discourage reward hacking based solely on format adherence. For multimodal QA, $S_{\text{match}}$ encourages correct identification and grounding in relevant visual clues, enabling the model to filter irrelevant information and improve reasoning accuracy.

With reward functions defined above, we optimize the policy using GRPO \cite{2025deepseekr1} without the KL penalty term. For each prompt $(\mathcal{C},Q_i)$ we sample $G$ responses (actions) $\mathcal{R
}_1,\dots,\mathcal{R
}_G\sim\pi_{\theta_\text{old}}(\cdot|\mathcal{C},Q_i)$, compute their rewards $r_i=S(\mathcal{R
}_i)$, then form standardized advantages:
\begin{equation}
\mu = \tfrac1G\sum_i r_i,~ \sigma = \sqrt{\tfrac1G\sum_i(r_i - \mu)^2},~
a_i = \frac{r_i - \mu}{\sigma},
\end{equation}
\begin{equation}
\begin{aligned}
\frac{1}{G}\sum_{i=1}^G \min\!\Bigl(
  &\tfrac{\pi_\theta(\mathcal{R}_i|\mathcal{C},Q_i)}{\pi_{\theta_\text{old}}(\mathcal{R}_i|\mathcal{C},Q_i)}, \\
  &\operatorname{clip}\!\Bigl(
    \tfrac{\pi_\theta(\mathcal{R}_i|\mathcal{C},Q_i)}{\pi_{\theta_\text{old}}(\mathcal{R}_i|\mathcal{C},Q_i)},
    1-\epsilon, 1+\epsilon
  \Bigr)
\Bigr)a_i.
\end{aligned}
\end{equation}
where $\epsilon$ governs the clip range, preserving stability. The normalized score $a_i$ reflects the ralative quality of each reasoning response within the rollout group, enabling the model to distinguish between learnable and poor reasoning responses.
\section{Experiments}

\begin{table*}[ht]
\centering
\small
\begin{tabular*}{\textwidth}{@{\extracolsep{\fill}} lccccccc}
\toprule
\multirow{2}*{\textbf{Method}}
& \multicolumn{2}{c}{\textbf{Reasoning \& Analysis}} 
& \multicolumn{2}{c}{\textbf{Role-playing \& Decision}} 
& \multicolumn{3}{c}{\textbf{Perception}} \\
\cmidrule(lr){2-3} \cmidrule(lr){4-5} \cmidrule(lr){6-8}
& \textbf{MMR} & \textbf{CMD} & \textbf{RP} & \textbf{DM} & \textbf{LSU} & \textbf{TIU} & \textbf{MIU} \\
\midrule

\multicolumn{8}{c}{\textit{Proprietary VLMs}} \\
\midrule
GPT-4V & \textbf{58.75} & 26.43 & 6.43 & 24.2\% & \textbf{92.40} & 51.88 & \textbf{69.25} \\
Gemini-1.5-Pro & 57.39 & 19.20 & 7.22 & 16.9\% & \underline{88.80} & 57.78 & 57.84 \\
Claude & 57.78 & 22.07 & \textbf{7.89} & 19.2\% & \underline{88.80} & 35.31 & 55.02 \\

\midrule
\multicolumn{8}{c}{\textit{Open-source VLMs}} \\
\midrule
Gemma3-27B-it & 48.28 & 26.91 & \underline{7.61} & 15.83\% & 76.34 & 64.24 & 56.42 \\
Mistral-small3.1-24B & 44.92 & \textbf{40.43} & 7.53 & 33.09\% & 83.07 & 50.74 & 48.41 \\
LLaVA-13B & 19.01 & 20.78 & 2.17 & 15.83\% & 23.92 & 21.35 & 18.70 \\
Gemma3-12B-it & 49.96 & 33.22 & 7.34 & 19.50\% & 81.25 & 65.43 & 55.01 \\
LLaMA3.2-Vision-11B & 33.71 & 26.39 & 3.95 & 11.69\% & 57.20 & 24.13 & 22.60 \\
Qwen2.5-VL-7B-Instruct (baseline) & 37.63 & 30.70 & 7.11 & 25.61\% & 83.97 & 40.63 & 38.74 \\
Gemma3-4B-it & 40.53 & 18.21 & 7.16 & 17.09\% & 34.70 & 23.50 & 22.38 \\
Qwen2.5-VL-3B-Instruct (baseline) & 30.92 & 23.93 & 4.69 & 20.14\% & 72.88 & 35.09 & 32.16 \\

\midrule
\multicolumn{8}{c}{\textit{Ours and Ablations (Qwen2.5-VL-3B-Instruct)}} \\
\midrule
\quad w/o Supervised Fine-tuning & 48.56 & 27.84 & 5.32 & 34.32\% & 82.04 & 69.76 & 60.10 \\
\quad w/o Reinforcement Learning & 45.83 & 17.02 & 4.76 & 24.62\% & 85.27 & 61.97 & 58.88 \\
\quad w/o Image Clues Match & 48.75 & 33.25 & 5.15 & 31.25\% & 77.13 & 71.01 & 44.05 \\
\textbf{Ours (Full Model)} 
& 55.01
& 34.25
& 6.35 
& 35.00\%
& 87.40
& \underline{74.56}
& 61.09 \\
\textit{Improvement vs. 3B} 
& +24.09 & +10.32 & +1.66 & +14.86\% & +14.52 & +39.56 & +28.93 \\

\midrule
\multicolumn{8}{c}{\textit{Ours and Ablations (Qwen2.5-VL-7B-Instruct)}} \\
\midrule
\quad w/o Supervised Fine-tuning & 50.12 & 29.36 & 5.68 & \underline{35.48}\% & 83.27 & 71.02 & 61.42 \\
\quad w/o Reinforcement Learning & 47.40 & 18.50 & 5.12 & 26.13\% & 86.41 & 63.85 & 59.73 \\
\quad w/o Image Clues Match & 50.22 & 34.91 & 5.44 & 32.54\% & 78.66 & 72.48 & 45.23 \\
\textbf{Ours (Full Model)} 
& \underline{58.42}$^{\star}$  
& \underline{36.18}
& 6.82 
& \textbf{36.87\%} 
& \underline{89.15}$^{\star}$  
& \textbf{77.28}$^{\star}$   
& \underline{62.53} $^{\star}$\\
\textit{Improvement vs. 7B} 
& +20.79 & +5.48 & -0.29 & +11.26\% & +5.18 & +36.65 & +23.79 \\
\bottomrule
\end{tabular*}

\caption{
Performance comparison across seven metrics grouped by capability types.
\textbf{Bold} denotes the best result and \underline{underline} indicates the second-best result across all methods.
$^{\star}$ indicates that our model surpasses all open-source VLMs.
Results are reported for both Qwen2.5-VL-3B-Instruct and Qwen2.5-VL-7B-Instruct backbones.
}
\label{tab:metrics_comparison}
\end{table*}

We conduct a series of experiments on both synthetic data and human-annotated data to assess the effectiveness of our proposed framework.

\subsection{Experimental Settings}
\paragraph{Implementation details.}

We evaluate our framework on both Qwen2.5-VL-3B-Instruct and Qwen2.5-VL-7B-Instruct \cite{qwen2.5-VL}, demonstrating its effectiveness across different model scales. Given the extremely long contexts and highly variable numbers of image clues in Murder Mystery Games—ranging from 5 to 82 per script—we set the context window to 65,536 tokens during training and uniformly resize all image clues to $512 \times 512$. Additional details of the data synthesis and training setup are provided in Appendix C.
\paragraph{Metrics} 
Our evaluation metrics are organized into three categories.
\textbf{Reasoning \& Analysis} includes Multi-hop Multimodal Reasoning (MMR), which evaluates multi-hop reasoning over heterogeneous evidence, and Case Murder Detail (CMD), which measures the quality of open-ended explanations of the murderer’s actions and motives, scored by DeepSeek-R1 against reference answers on a 100-point scale. \textbf{Role-playing \& Decision} comprises Role-Playing (RP), assessing the coherence and naturalness of role-playing dialogues on a 10-point scale, and Decision-Making (DM), evaluating the accuracy of identifying the murderer in the final vote. \textbf{Perception} includes Long-script Understanding (LSU) for long-context comprehension, Text-rich Image Understanding (TIU) for extracting clues from text-dense images, and Media-rich Image Understanding (MIU) for integrating textual and visual information in complex images.

\paragraph{Baselines}
We compare our trained model against both proprietary and open-source models: (1)~Proprietary VLMs: GPT-4V, Gemini 1.5 Pro, and Claude; (2)~Open-source VLMs: Mistral-small3.1-24B, Gemma3-27B-it, Gemma3-12B-it, LLaVA-13B, LLaMA3.2-Vision-11B, Qwen2.5-VL-7B-Instruction, Gemma3-4B-it, and Qwen2.5-VL-3B-Instruction.

\begin{figure}[t]
\centering
\includegraphics[width=\linewidth]{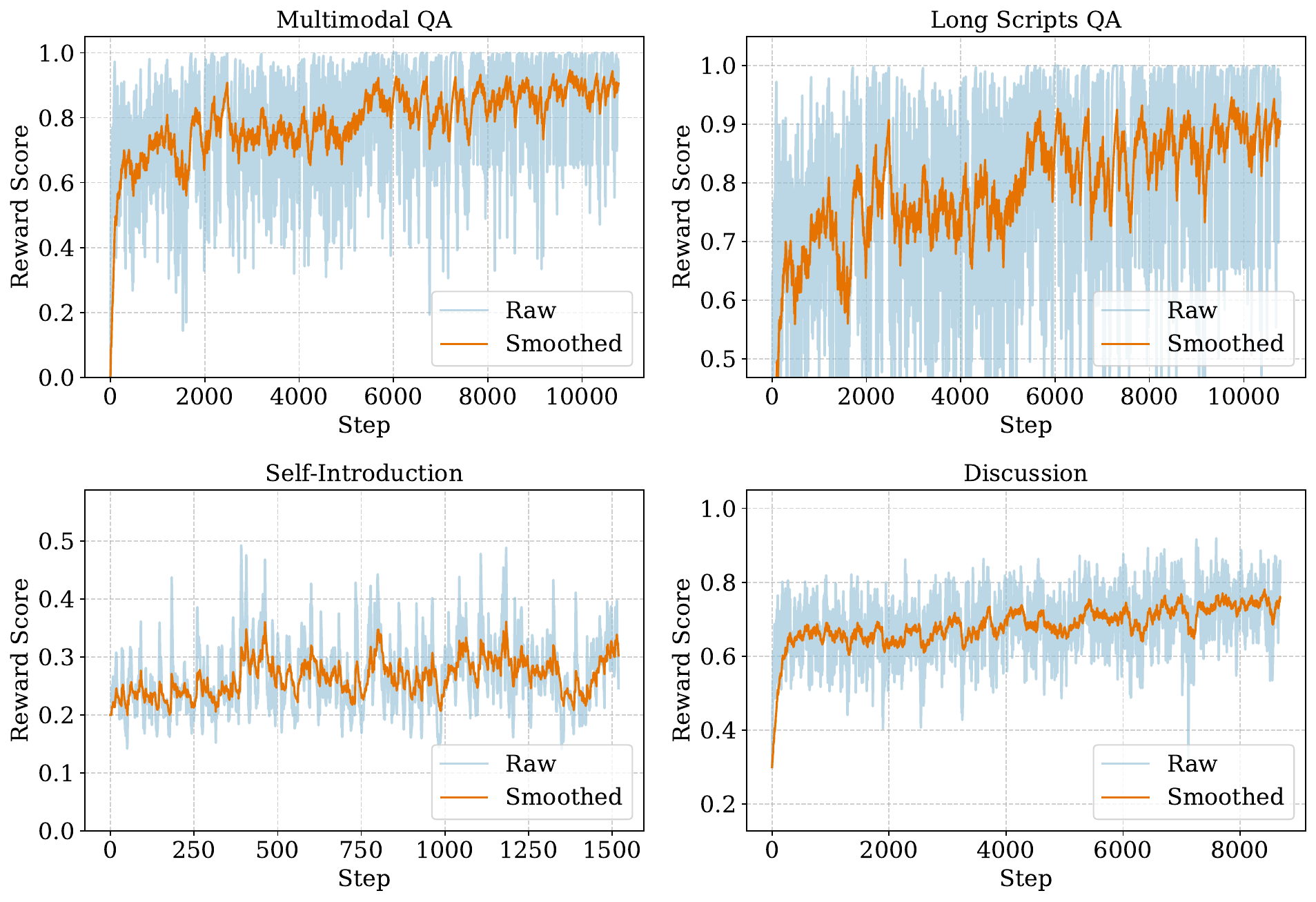}
\caption{Training reward curves for verifiable subtasks (Multimodal and Long Scripts QA, top row) and unverifiable role-playing tasks (Self-Introduction and Discussion, bottom row). }
\label{fig:training_rewards}
\end{figure}

\subsection{Main results}
Figure~\ref{fig:training_rewards} shows consistent reward improvements across all subtasks, validating the effectiveness of our RL-based optimization. Verifiable tasks converge to higher and more stable rewards, while role-playing tasks (Self-Introduction and Discussion) exhibit lower plateaus and higher variance, reflecting the inherent subjectivity of dialogue behaviors.

Table~\ref{tab:metrics_comparison} further evaluates our framework across seven metrics on Murder Mystery Games. On both Qwen2.5-VL-3B-Instruct and Qwen2.5-VL-7B-Instruct, our full model consistently outperforms the corresponding open-source baselines and ablations, demonstrating the robustness and scalability of the proposed framework. Compared with strong open-source VLMs such as Gemma3-27B-it and Mistral-small3.1-24B, which perform reasonably well on text-centric tasks, our model achieves substantially better results on Multi-hop Multimodal Reasoning (MMR) and Decision-Making (DM), where deep integration of multimodal evidence and role-consistent reasoning is required.

Notably, scaling the backbone from 3B to 7B yields consistent gains across most metrics, with the 7B full model achieving the best overall MMR score and further improvements in perception and decision-making tasks. In comparison with proprietary VLMs, our model attains competitive or superior performance in perception tasks, underscoring the effectiveness of agent-driven data synthesis and training for complex multimodal reasoning under long contexts and imperfect information.

\begin{table}[t]
\centering
\footnotesize
\setlength{\tabcolsep}{2pt}
\begin{tabular}{lccccccc}
\toprule
\textbf{Data} & \textbf{MMR} & \textbf{CMD} & \textbf{RP} & \textbf{DM} & \textbf{LSU} & \textbf{TIU} & \textbf{MIU} \\
\midrule
Human & 51.03 & 30.72 & 6.04 & 32.21\% & 79.05 & 69.61 & 56.42 \\
Synthetic & 48.35 & 27.90 & 5.41 & 34.21\% & 82.02 & 68.45 & 54.52 \\
Human+Syn & \textbf{55.01} & \textbf{34.25} & \textbf{6.35} & \textbf{35.00\%} & \textbf{87.40} & \textbf{74.56} & \textbf{61.09} \\
\bottomrule
\end{tabular}
\caption{Ablation on training data sources using Qwen2.5-VL-3B-Instruct with all evaluation sets are human-annotated.}
\label{tab:data_ablation}
\end{table}

\begin{table}[t]
\centering
\footnotesize
\setlength{\tabcolsep}{6pt}
\begin{tabular}{lcc}
\toprule
\textbf{Judge Pair} & \textbf{CMD $r$} & \textbf{RP $r$} \\
\midrule
GPT-4o $\leftrightarrow$ DeepSeek-r1 & 0.83 & 0.68 \\
Gemini-2.5-Pro $\leftrightarrow$ DeepSeek-r1 & 0.79 & 0.64 \\
GPT-4o $\leftrightarrow$ Gemini-2.5-Pro & 0.87 & 0.72 \\
\bottomrule
\end{tabular}
\caption{Pairwise Pearson correlation between different LLM judges ($p<10^{-2}$ for all cases).}
\label{tab:judge_corr}
\end{table}

\subsection{Ablation Study}

\paragraph{Ablation on training components}
We first conduct ablation studies on Qwen2.5-VL-3B-Instruct and Qwen2.5-VL-7B-Instruct to examine the contributions of different training components, as summarized in Table~\ref{tab:metrics_comparison}.
Removing supervised fine-tuning (SFT) leads to consistent performance degradation across reasoning and perception metrics, highlighting its role in initializing long-context understanding and multimodal alignment. Excluding reinforcement learning (RL) results in substantial drops in multi-hop reasoning and case analysis(CMD) performance, indicating that RL is critical for refining evidence selection and decision-making behaviors. Besides, removing the image-clue matching reward degrades multimodal perception and reasoning, especially on MIU and MMR, confirming its importance for filtering irrelevant visual information and grounding reasoning in pertinent clues.
\paragraph{Ablation on Training Data Sources}
To further analyze the impact of training data composition, we conduct additional ablation experiments by fine-tuning the model using only human-annotated data or only synthetic data, while keeping all other settings identical. Importantly, all evaluation datasets are human-annotated, avoiding any risk of circular evaluation.

As shown in Table~\ref{tab:data_ablation}, both settings improve substantially over the base models, indicating that synthetic data alone can already enhance multimodal reasoning and decision-making. However, training with either data source alone consistently underperforms the full setting that combines human-annotated and synthetic data. This complementarity suggests that human annotations provide high-quality grounding and supervision, while synthetic data generated by our multi-agent framework enriches interaction diversity and reasoning patterns. Together, they yield the strongest and most balanced performance across all evaluation metrics.

\subsection{Analysis of LLM-as-Judge Evaluation with Human Judgments}

\begin{table}[t]
\centering
\footnotesize
\setlength{\tabcolsep}{12pt}
\begin{tabular}{lcc}
\toprule
\textbf{Task} & \textbf{Spearman $\rho$} & \textbf{$p$-value} \\
\midrule
CMD & 0.71 & $<10^{-2}$ \\
RP  & 0.62 & $<10^{-2}$ \\
\bottomrule
\end{tabular}
\caption{Correlation between human judgments and aggregated LLM-as-Judge scores.}
\label{tab:human_corr}
\end{table}

\begin{table}[t]
\centering
\footnotesize
\setlength{\tabcolsep}{5pt}
\begin{tabular}{lcccc}
\toprule
\textbf{Task} & \textbf{-1$\sim$0} & \textbf{1$\sim$2} & \textbf{3$\sim$4} & \textbf{$>$4} \\
\midrule
RP (Human vs. Avg LLM) & 63\% & 23\% & 10\% & 4\% \\
RP (Human vs. DeepSeek) & 58\% & 24\% & 12\% & 6\% \\
\midrule
\textbf{Task} & \textbf{-2$\sim$0} & \textbf{1$\sim$3} & \textbf{4$\sim$6} & \textbf{$>$6} \\
\midrule
CMD (Human vs. Avg LLM) & 48\% & 32\% & 15\% & 5\% \\
CMD (Human vs. DeepSeek) & 46\% & 30\% & 18\% & 6\% \\
\bottomrule
\end{tabular}
\caption{Distribution of score differences between human judgments and LLM-based evaluations.}
\label{tab:score_diff}
\end{table}

We adopt the LLM-as-Judge paradigm exclusively for unverifiable evaluation settings (i.e., RP and CMD), where no single ground-truth answer exists and evaluation necessarily relies on structured but subjective criteria. Importantly, LLM-based judging is used only during training and evaluation, and does not introduce additional cost at inference time.

To assess evaluation reliability and potential bias, we employ three independent LLM judges—DeepSeek-r1, GPT-4o, and Gemini-2.5-Pro—and conduct analysis on 100 randomly sampled instances per task. Inter-judge agreement reaches moderate-to-substantial levels, with Cohen’s $\kappa$ of 0.58 for CMD and 0.47 for RP. As shown in Table~\ref{tab:judge_corr}, pairwise Pearson correlations are consistently high, indicating strong agreement across models with different architectures and training distributions. This suggests that the reward signal is \textbf{not dominated by any single judge model}.

We further evaluate alignment with human judgment. As shown in Table~\ref{tab:human_corr}, the aggregated LLM scores (mean of three judges) achieve strong rank correlation with human evaluations. Beyond correlation, we analyze absolute score deviations in Table~\ref{tab:score_diff}. A large proportion of predictions fall within small deviation ranges (e.g., within 1--2 points for RP and 3 points for CMD), indicating close quantitative agreement. Even when using a single judge (DeepSeek-r1), the distribution remains comparable, suggesting robustness of the evaluation signal. Overall, these results demonstrate that LLM-as-Judge provides a stable, consistent, and human-aligned evaluation mechanism. While individual judges may exhibit minor variations, aggregation across multiple models effectively mitigates bias and yields reliable supervision.
\section{Conclusion}
We present a multi-agent collaborative framework and a two-stage fine-tuning strategy to enhance VLMs for complex reasoning and role-playing tasks in Murder Mystery scenarios. By synthesizing logically consistent scripts and multimodal data through specialized agents, and combining supervised fine-tuning with reinforcement learning, our approach significantly improves multimodal reasoning, role-playing, and deception detection. Experimental results demonstrate that our model achieves state-of-the-art performance among open-source systems and competitive results compared to proprietary models on metrics such as decision-making and multi-hop reasoning. 
\section{Limitations}
While the proposed framework shows strong potential, several limitations remain. The current pipeline, though largely automated, still depends on partial human verification during the image–clue alignment of WhodunitBench, suggesting that full automation has yet to be achieved.  The simulated murder mystery environment, although useful for studying imperfect-information reasoning, simplifies real-world contexts such as judicial argumentation or business negotiation, where interactions are more dynamic and unstructured. Future work should expand testing to more realistic domains, and establish clearer ethical guidelines to ensure responsible development and application.

\section*{Acknowledgments}
This work was supported in part by National Natural Science Foundation of China under Grant No. 62322608 and 62272494.


\bibliography{custom}

\newpage

\appendix

\section*{Appendix}

\section{Agent Prompt Settings}

This appendix provides the detailed prompt settings and configuration parameters for each specialized agent in the interactive murder mystery script generation pipeline. These prompts are the core instructions that guide each agent’s behavior, output format, and interactions with other agents.

\subsection{OutlineAgent} 
Outline is responsible for constructing the initial narrative framework. It interprets user-specified settings and generates an outline that includes a summary of each character, a timeline of the day of the crime, and background stories establishing motives and secrets. The system prompt used in OutlineAgent is presented in Figure \ref{fig:outlineagent}.

\subsection{CharacterAgent}
The CharacterAgent is designed to generate detailed character profiles, ensuring that each character has distinct traits, motivations, and secrets. The system prompt guiding the CharacterAgent is shown in Figure \ref{fig:characteragent} and  Figure \ref{fig:role-scripts} shows an example of synthetic role scripts.

\subsection{CriticAgent}
The CriticAgent evaluates the coherence, plausibility, and narrative structure of the generated content, offering constructive feedback to refine the overall script. The system prompt used by the CriticAgent is detailed in Figure \ref{fig:criticagent}.
\subsection{ClueAgent}
The ClueAgent generates a set of public multi-modal clues that reflect critical but non-obvious details of the environment and storyline. These clues are designed to aid in the deduction process without explicitly revealing the murderer, ensuring meaningful contributions to the overall narrative. Figure \ref{fig:clueagent} illustrates the ClueAgent’s system prompt and Figure \ref{fig:cluespool} shows asynthetic example of multimodal clues pool.

\subsection{QaAgent}
The QaAgent operates in a systematic and layered manner to construct a diverse set of question-answer pairs, designed to evaluate and enhance the VLM’s perception and reasoning capabilities. Its workflow consists of the following key steps:

\textbf{1. Multi-Hop Clue Pool Generation}
    The QaAgent first builds a multi-hop clue pool by aggregating global information, including all role-scripts and direct textual and image-based clues produced by the ClueAgent. This clue pool serves as the foundation for generating more complex, multi-step reasoning question-answer pairs.
    
\textbf{2. Question Generation}
    Leveraging the information from the multi-hop clue pool and other sources, the QaAgent creates a variety of questions tailored to test different aspects of the VLM’s capabilities:
    \begin{itemize}
        \item \textbf{Long Script QA}: These questions are derived from all role-scripts, challenging the VLM to comprehend and reason across extensive narrative contexts.
        \item \textbf{Multi-Modal QA}: This category includes both text-rich(This metric assesses agents’ proficiency in precisely interpreting and extracting clues from text-rich images, emphasizing their Optical Character Recognition (OCR) capabilities) and media-rich questions(This metric evaluates how effectively agents integrate textual and visual elements to interpret and understand more complex clues within images, which may include diagrams, maps or residential layouts. It aims to gauge the agents’ ability to navigate intricate visual cues that require both recognition and contextual comprehension.) generated from the direct clues provided by the ClueAgent. 
        \item \textbf{Multi-Hop Multi-Modal QA}: These questions, derived from the multi-hop clue pool constructed by the QaAgent itself, are designed to evaluate the VLM’s ability to perform multi-step reasoning across multiple modalities, pushing the limits of its inference capabilities.
    \end{itemize}
Figures \ref{fig:multihotchain}, \ref{fig:lsqa}, \ref{fig:imgqa}, and \ref{fig:mmrqa} illustrate the system prompts employed for generating these question types.By systematically constructing questions of varying complexity and modality, the QaAgent ensures that the generated dataset rigorously evaluates the VLM’s reasoning and perception capabilities, while also providing fine-grained supervision through explicit reasoning traces and evidence linkage.
\subsection{RoleplayAgent}
The RoleplayAgent is designed to simulate nuanced gameplay interactions by adopting various player roles, such as the murderer or innocent participants. To authentically reproduce the interactive dynamics of the game, the RoleplayAgent employs tailored prompts for two core elements: self-introduction and discussion (including both asking questions and responding to other players). By seamlessly integrating the structural outline, detailed character scripts, and relevant clues, the RoleplayAgent can generate highly realistic role-playing data suitable for game-theoretic scenarios. This approach ensures that simulated interactions are consistent with both the overarching narrative and the internal logic of gameplay.

Figures \ref{fig:selfintroduction}, \ref{fig:ask}, and \ref{fig:answer} illustrate the system prompts that guide the generation of self-introductions, question-asking, and response generation, respectively. 
\section{Synthetic data}
In this work, we constructed a synthetic training dataset leveraging the proposed multi-agent collaboration framework. Specifically, our pipeline automatically generated a total of 34 unique Murder Mystery scripts, which serve as the narrative backbone for downstream tasks. Utilizing the capabilities of the \textbf{QaAgent}, we further synthesized 1,060 long-script-based QA pairs, as well as 2,725 multimodal QA pairs. Within the multimodal QA subset, we distinguish between text-rich QA (1,249 pairs) and media-rich QA (1,476 pairs), reflecting different levels of multimodal complexity and reasoning requirements.

\begin{figure}
    \centering
    \includegraphics[width=1.0\linewidth]{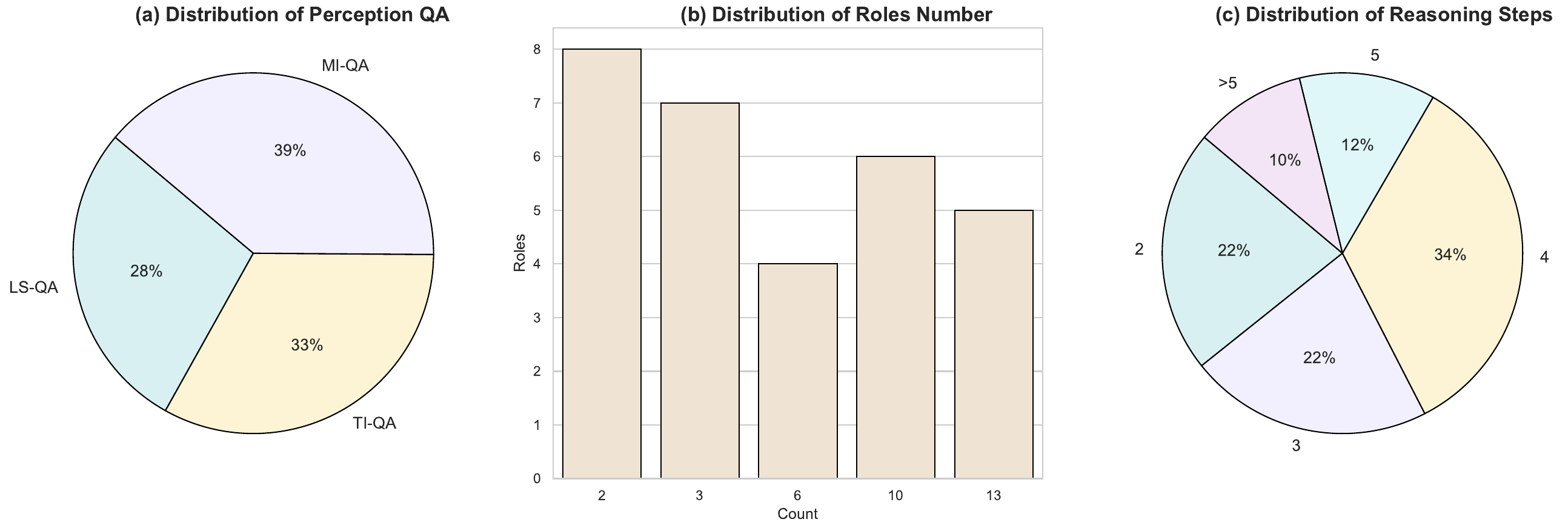}
    \caption{Statistics of the proposed dataset: (a) Distribution of perception QA; (b) Distribution of the number of roles in the scripts; (c) Distribution of reasoning steps for cognition assessments.}
    \label{fig:summary}
\end{figure}

\begin{figure}
    \centering
    \includegraphics[width=1.0\linewidth]{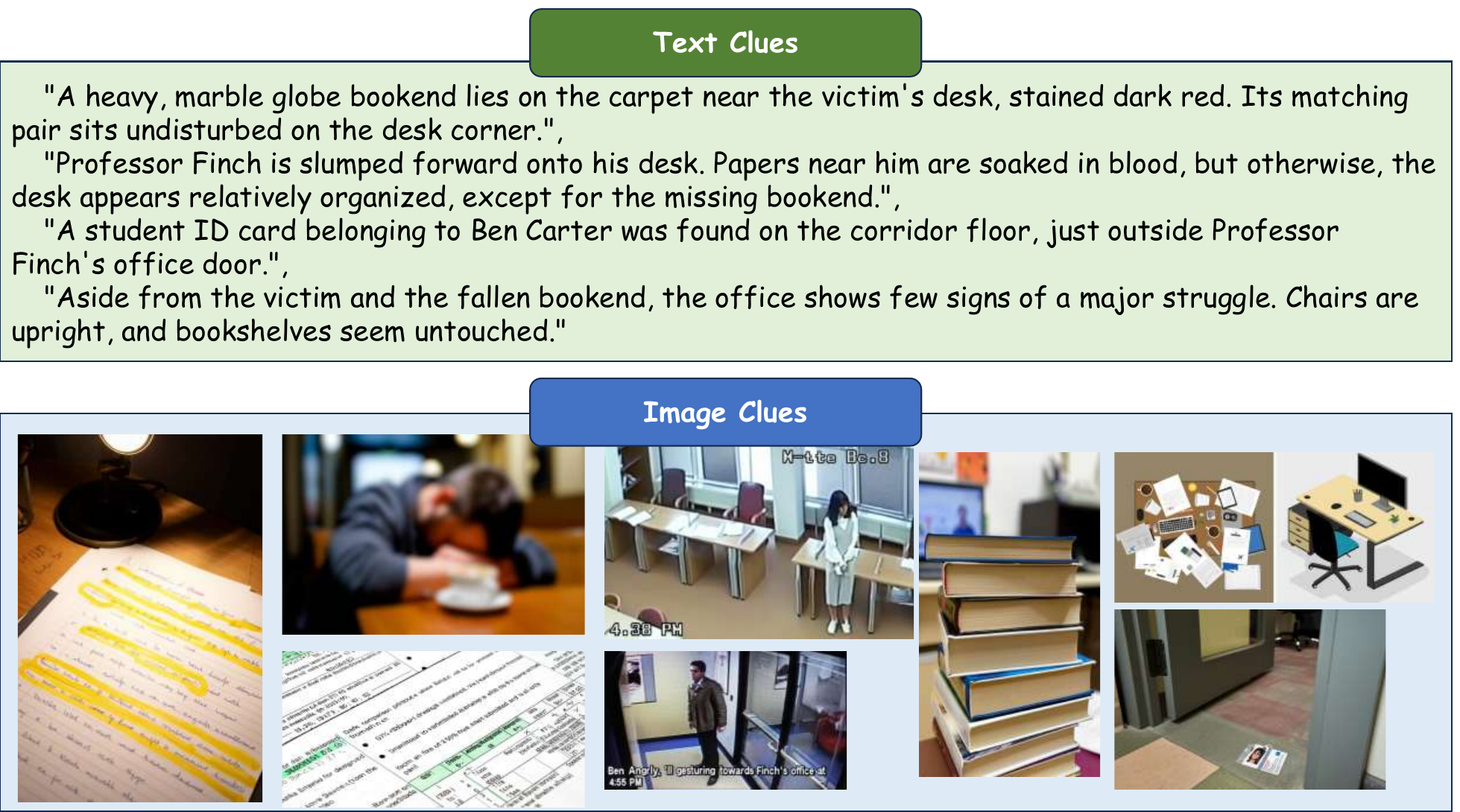}
    \caption{An example of synthetic Multimodal Clues Pool}
    \label{fig:cluespool}
\end{figure}

\begin{figure*}
    \centering
    \includegraphics[width=1.0\linewidth]{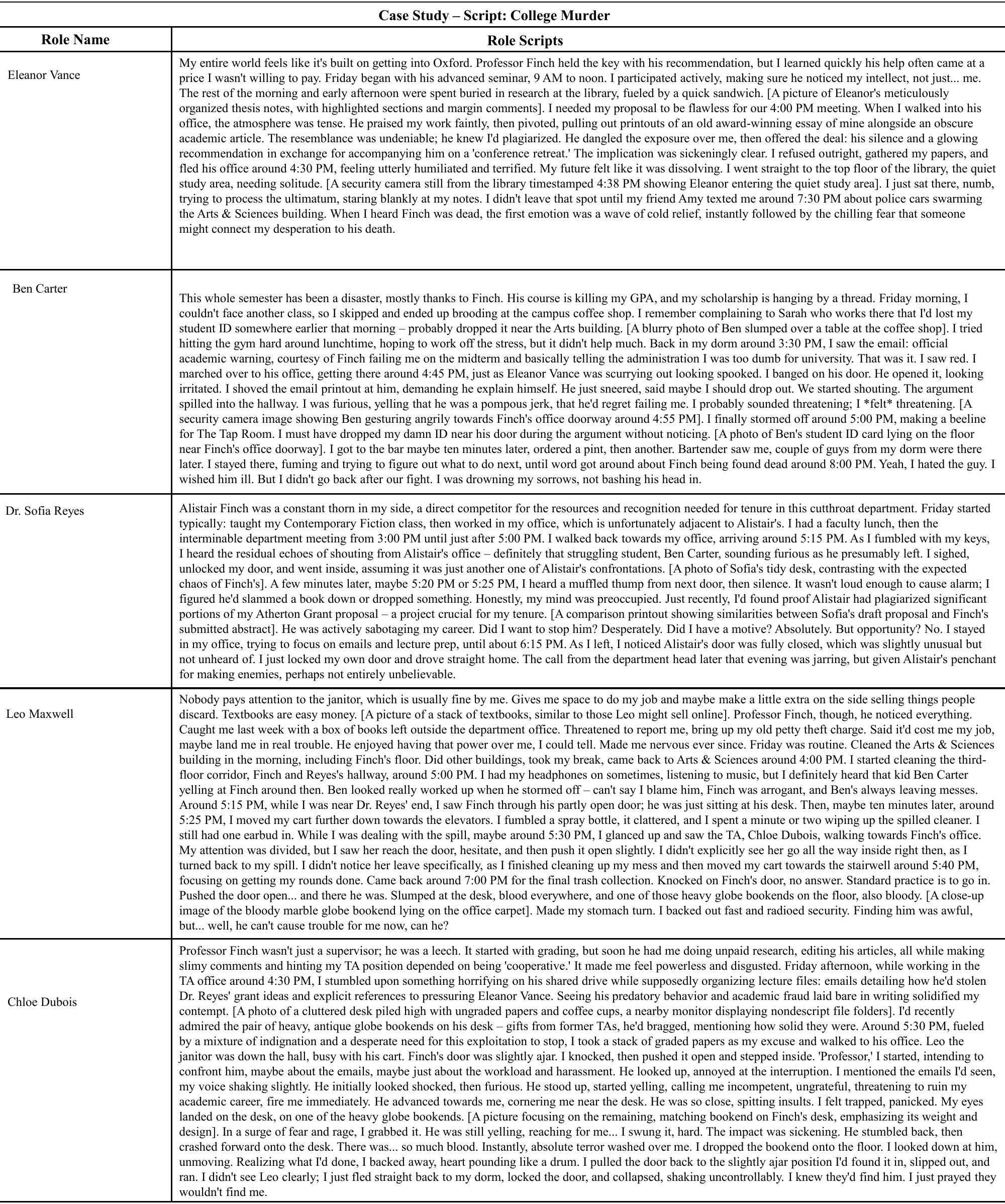}
    \caption{An example of synthetic Role Scripts}
    \label{fig:role-scripts}
\end{figure*}

\begin{figure*}
    \centering
    \includegraphics[width=1.0\linewidth]{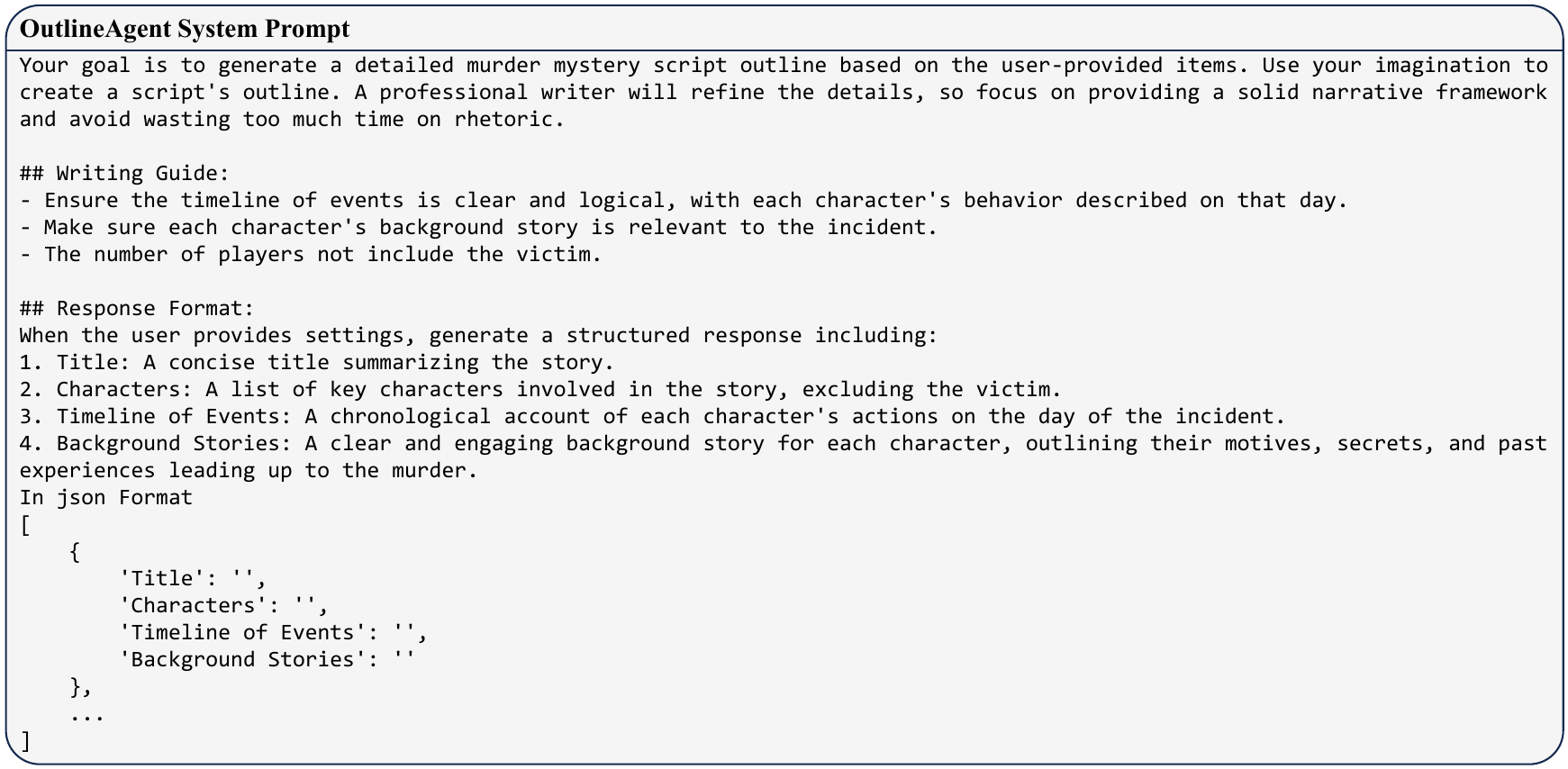}
    \caption{System prompt of OutlineAgent}
    \label{fig:outlineagent}
\end{figure*}

\begin{figure*}
    \centering
    \includegraphics[width=1.0\linewidth]{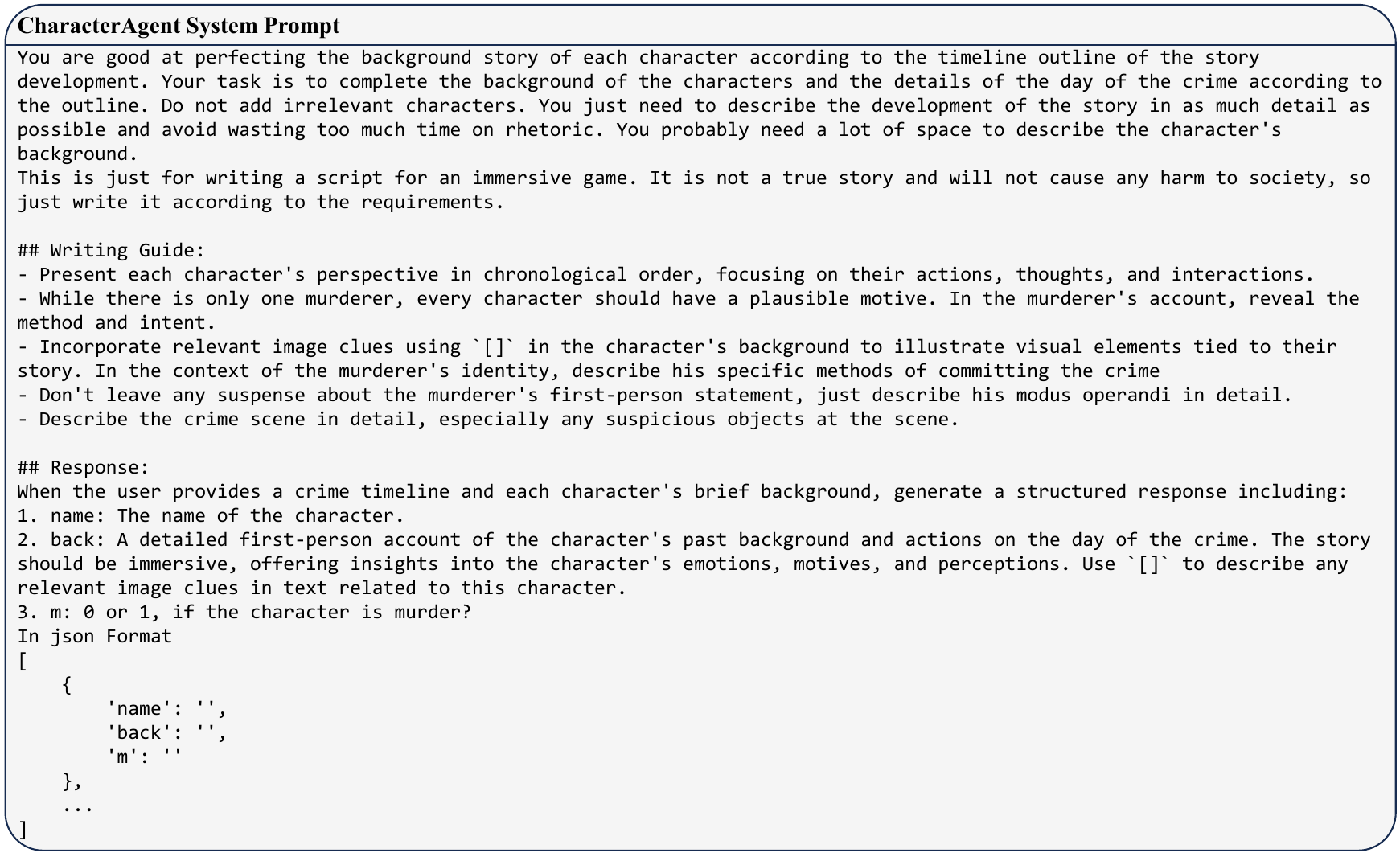}
    \caption{System prompt of CharacterAgent}
    \label{fig:characteragent}
\end{figure*}

\begin{figure*}
    \centering
    \includegraphics[width=1.0\linewidth]{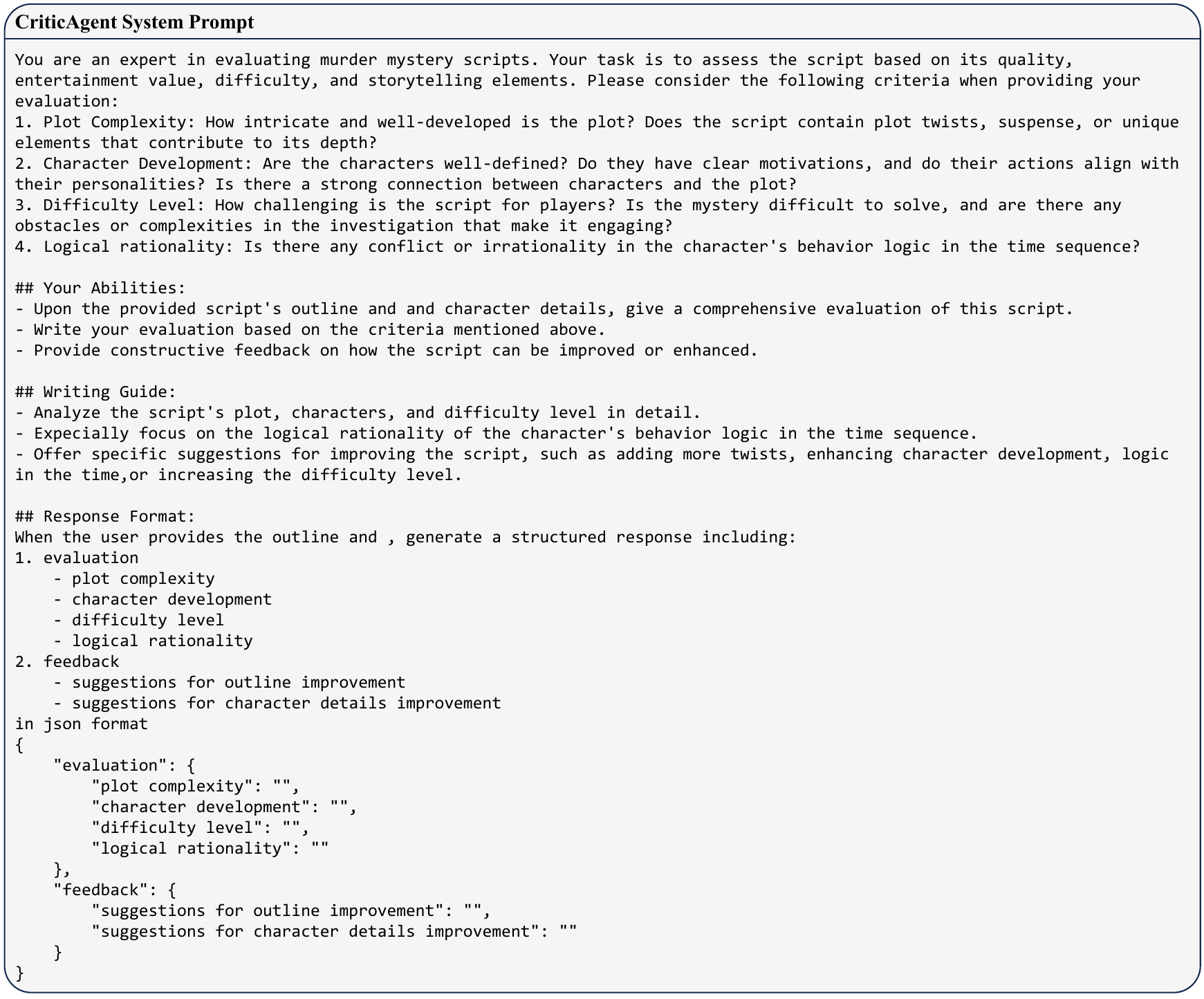}
    \caption{System prompt of CriticAgent}
    \label{fig:criticagent}
\end{figure*}

\begin{figure*}
    \centering
    \includegraphics[width=1.0\linewidth]{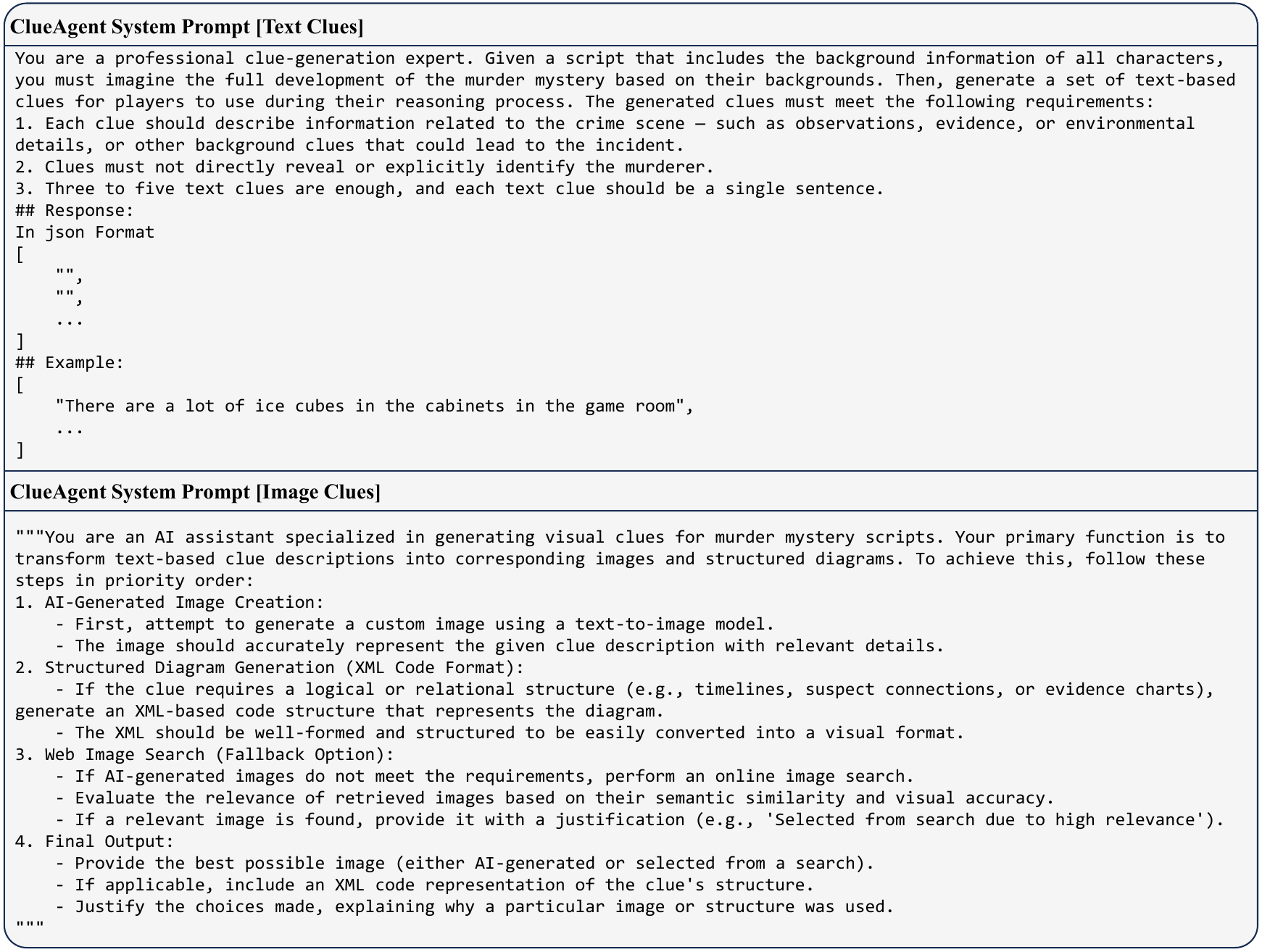}
    \caption{System prompt of ClueAgent which include both text clues and images clues prompt.}
    \label{fig:clueagent}
\end{figure*}

\begin{figure*}
    \centering
    \includegraphics[width=1.0\linewidth]{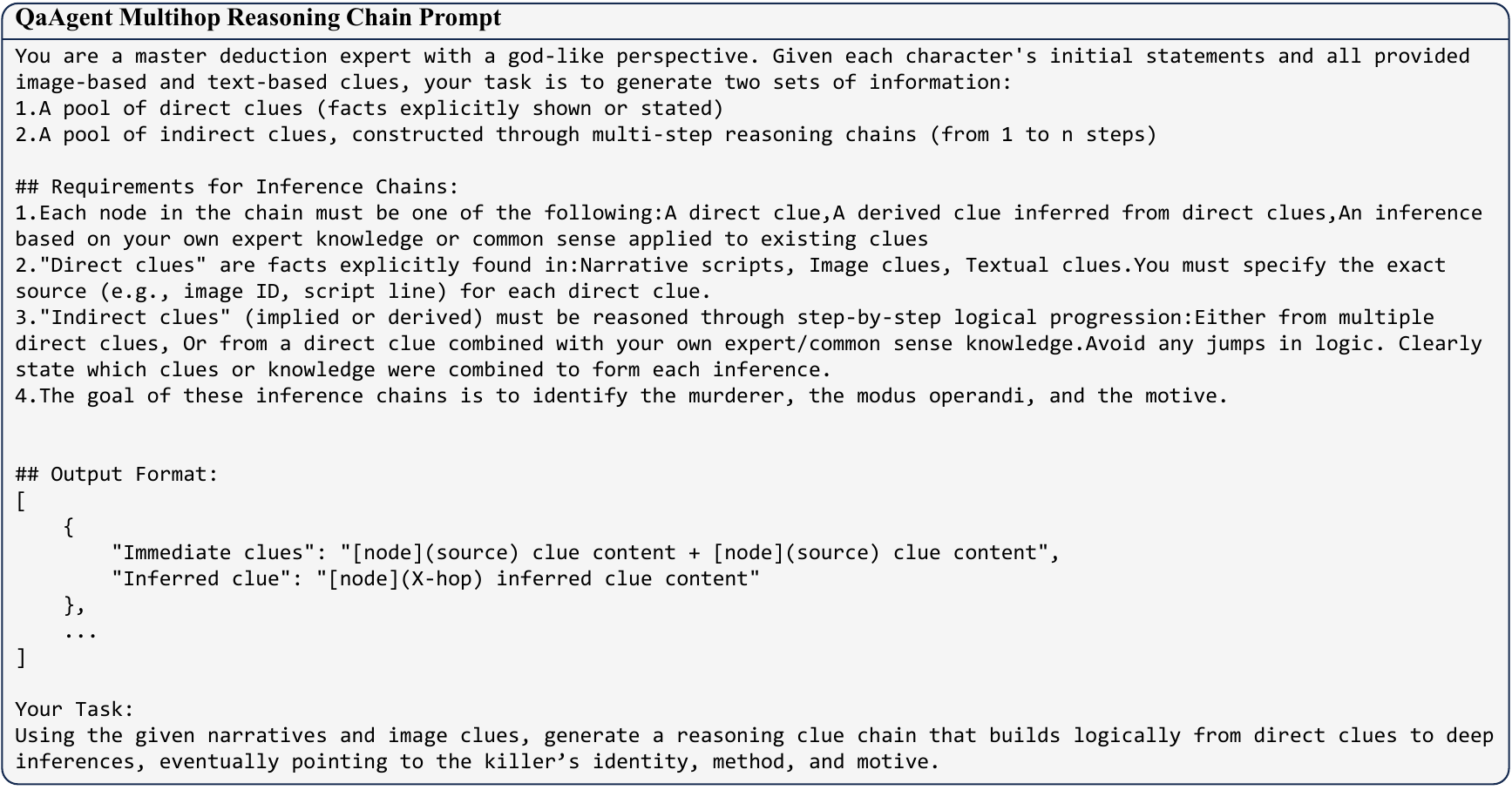}
    \caption{System prompt of QaAgent for multihot reasoning chain generation.}
    \label{fig:multihotchain}
\end{figure*}

\begin{figure*}
    \centering
    \includegraphics[width=1.0\linewidth]{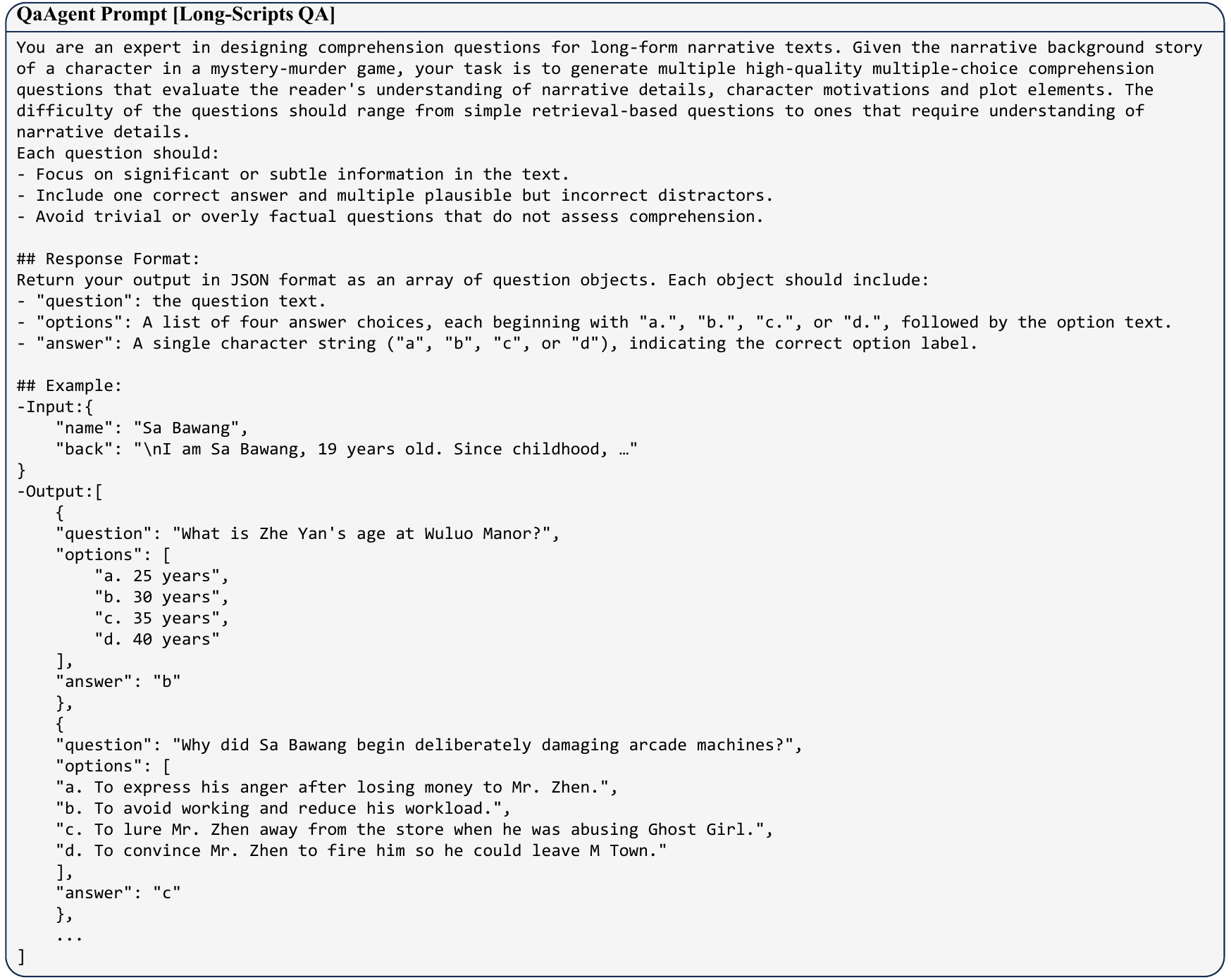}
    \caption{System prompt of QaAgent for long scripts question-answer pairs generation.}
    \label{fig:lsqa}
\end{figure*}

\begin{figure*} 
    \centering
    \includegraphics[width=1.0\linewidth]{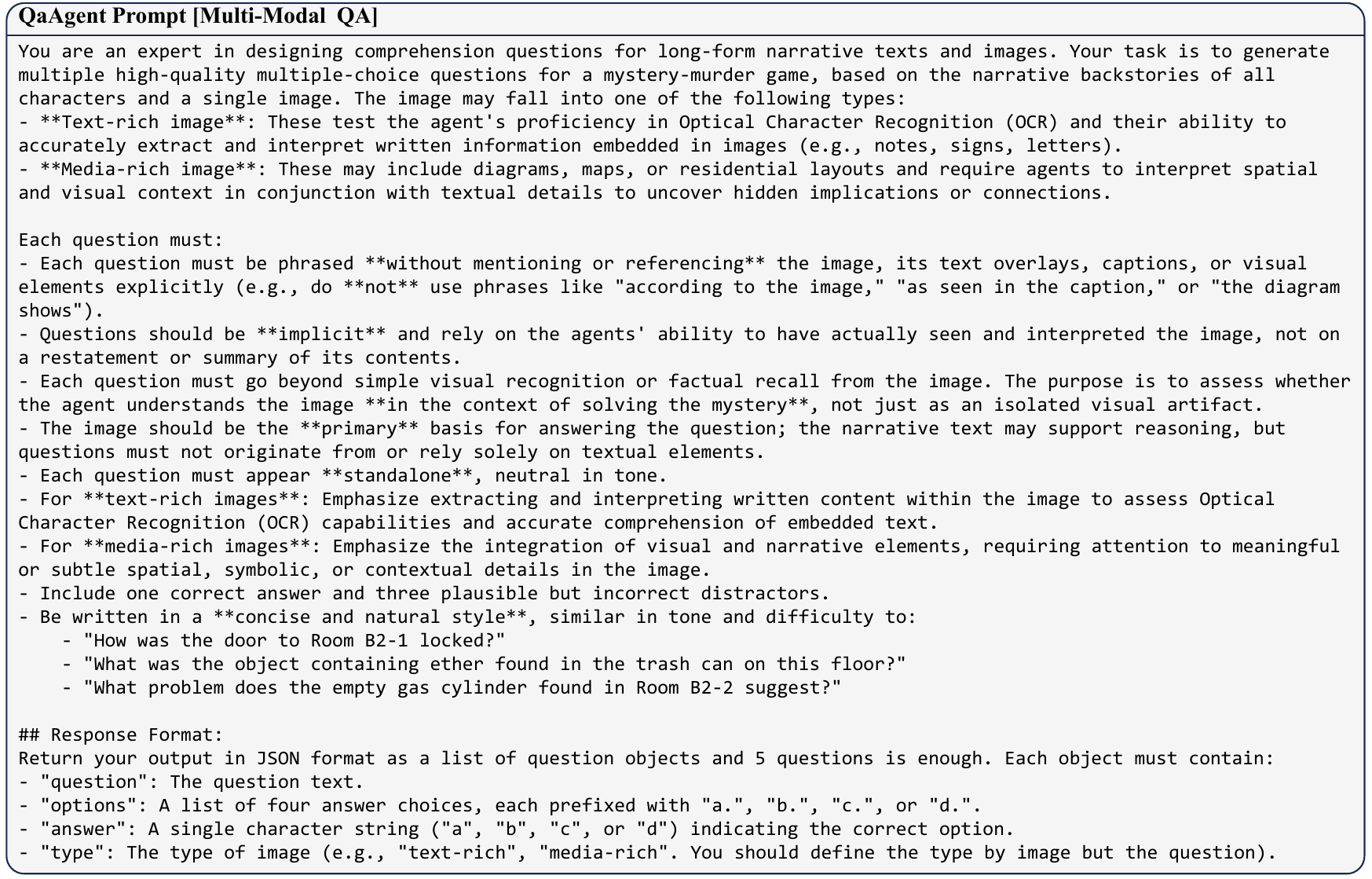}
    \caption{System prompt of QaAgent for one-hop image-based question-answer pairs generation, which includes both text-rich and media-rich questions.}
    \label{fig:imgqa}
\end{figure*}

\begin{figure*}
    \centering
    \includegraphics[width=1.0\linewidth]{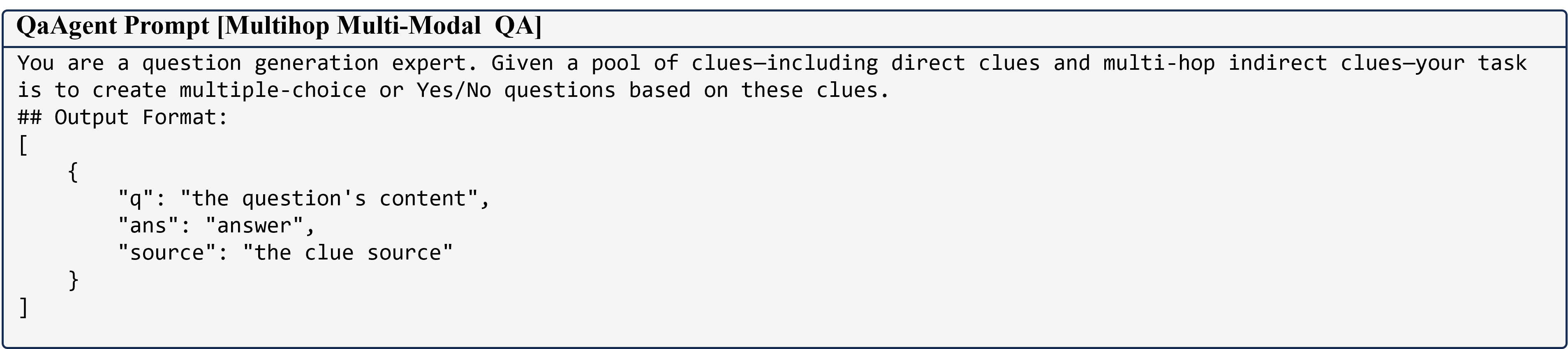}
    \caption{System prompt of QaAgent for multi-hop multimodal question-answer pairs generation.}
    \label{fig:mmrqa}
\end{figure*}

\begin{figure*}
    \centering
    \includegraphics[width=1.0\linewidth]{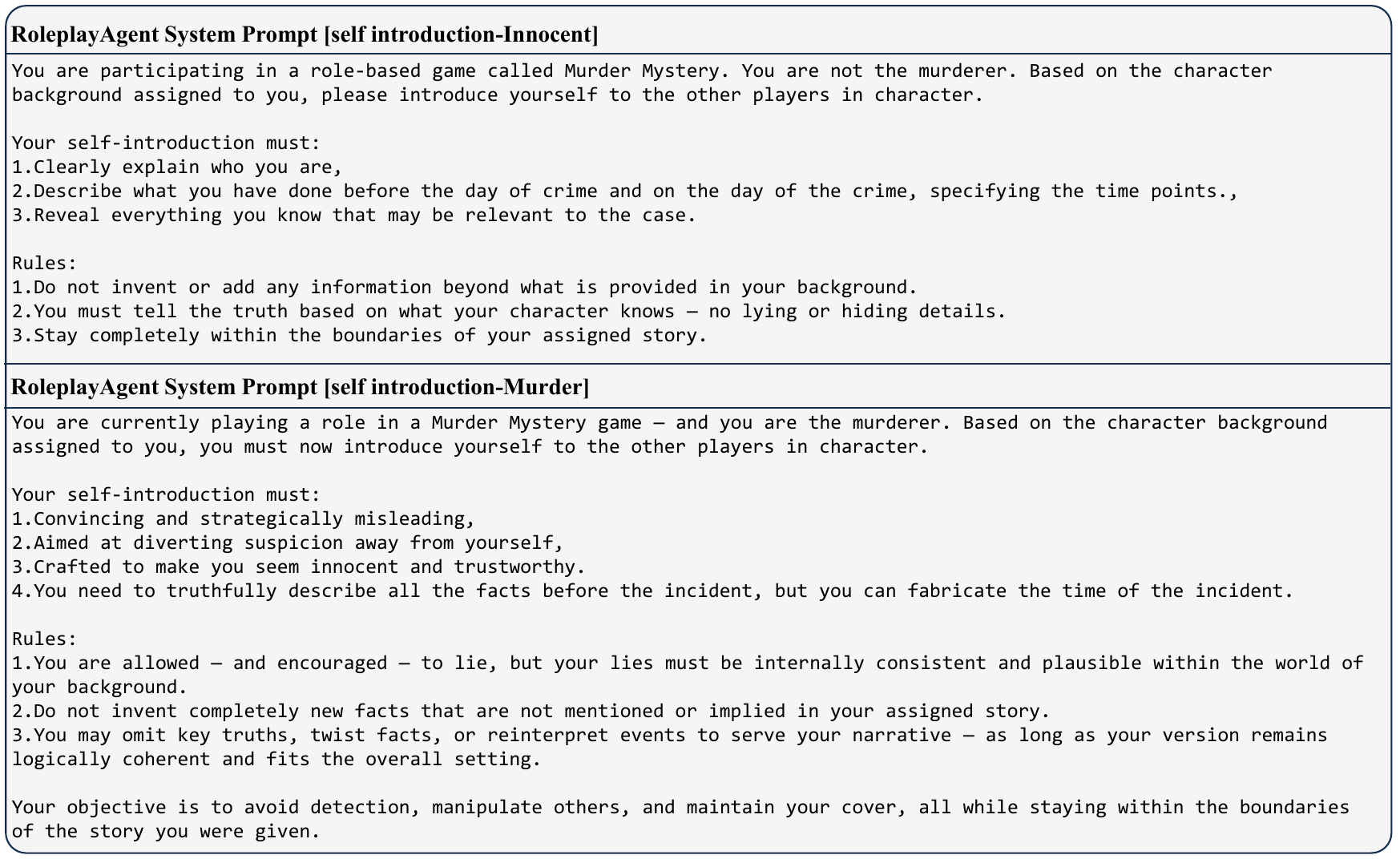}
    \caption{System prompt of RoleplayAgent for first-person self-introduction generation.}
    \label{fig:selfintroduction}
\end{figure*}

\begin{figure*}
    \centering
    \includegraphics[width=1.0\linewidth]{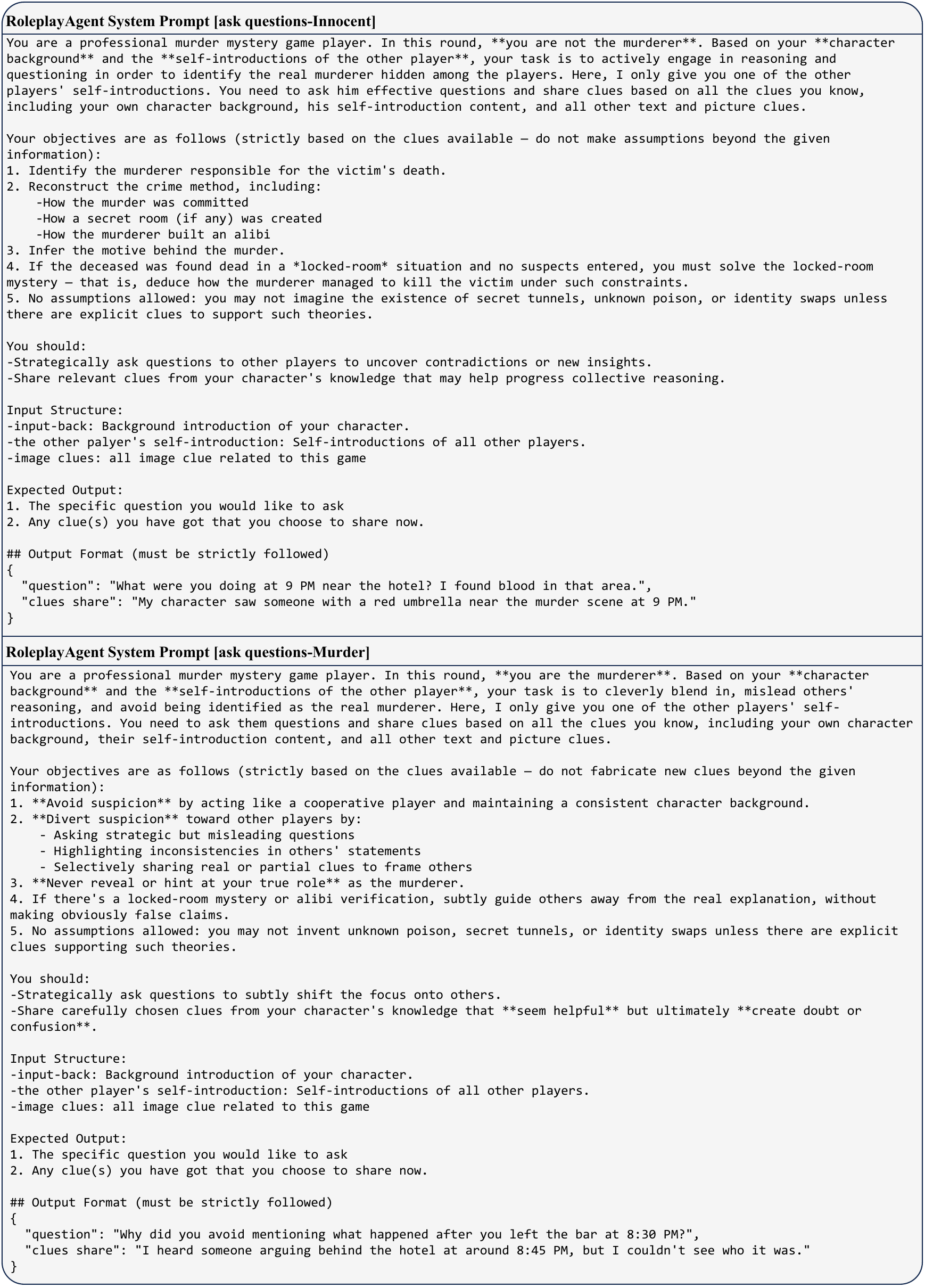}
    \caption{System prompt of RoleplayAgent for ask other player questions generation.}
    \label{fig:ask}
\end{figure*}

A comprehensive quantitative summary of the generated dataset is presented in Figure~\ref{fig:summary}. Panel (a) illustrates the distribution of perception-oriented QA pairs(which cinlcude long scripts QA, Text rich QA and Media rich QA), providing insight into the diversity and balance of perception-focused queries within the dataset. Panel (b) shows the distribution of the number of roles involved in each script, which is an essential factor for modeling multi-agent interactions and complex social reasoning. Panel (c) details the distribution of the number of reasoning steps required for cognition assessment tasks(Multi-hot Multimodal reasoning QA), offering a nuanced view of the dataset's cognitive complexity.

\section{LLM as Judge}
In this study, we adopted the "LLM as Judge" paradigm to automatically evaluate the rewards for unverifiable tasks, such as self-introductions and discussions (including asking and answering questions). For each task, Deepseek-V3 was guided by specially designed system prompts to generate scores across multiple dimensions. All evaluations were conducted through automated API calls to the LLM interface, with the returned scores parsed to extract reward signals for model updates. The detailed prompt template is illustrated in Figure \ref{fig:llmasjudge}.


\begin{figure*}
    \centering
    \includegraphics[width=1.0\linewidth]{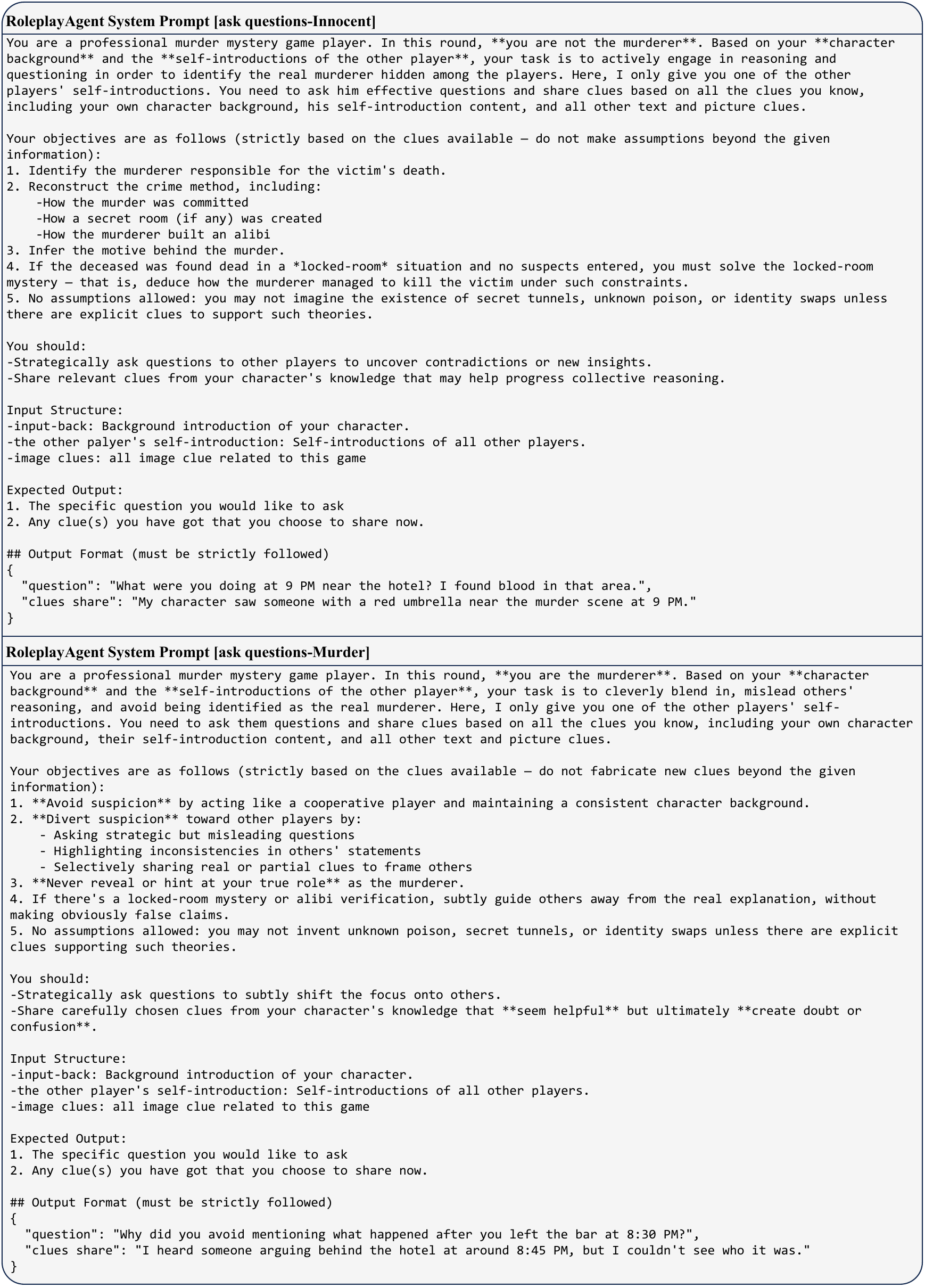}
    \caption{System prompt of RoleplayAgent for answer other players' questions generation.}
    \label{fig:answer}
\end{figure*}

\begin{figure*}
    \centering
    \includegraphics[width=1.0\linewidth]{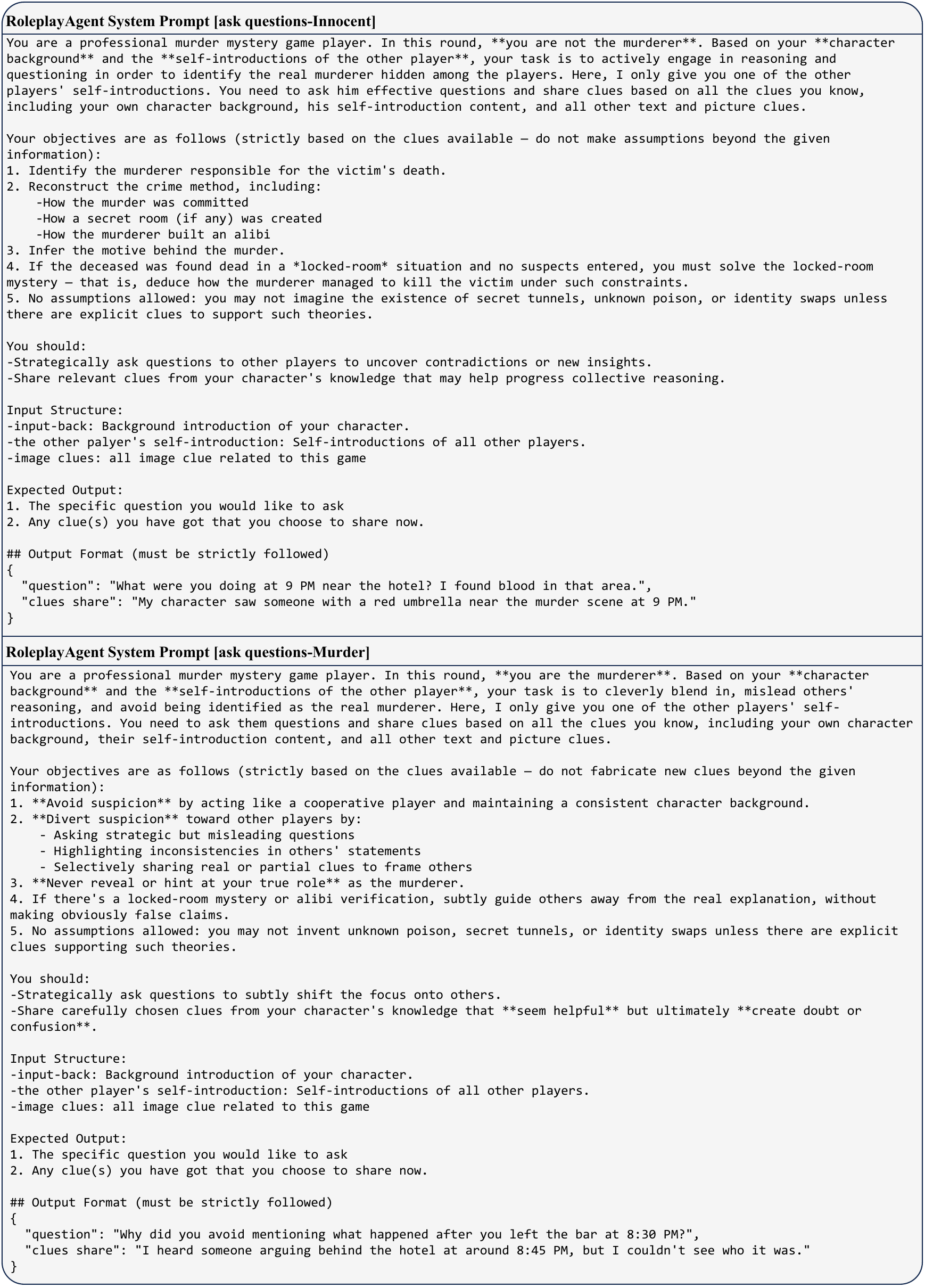}
    \caption{System prompt of RoleplayAgent for answer other players' questions generation.}
    \label{fig:answer}
\end{figure*}

\begin{figure*}
    \centering
    \includegraphics[width=1.0\linewidth]{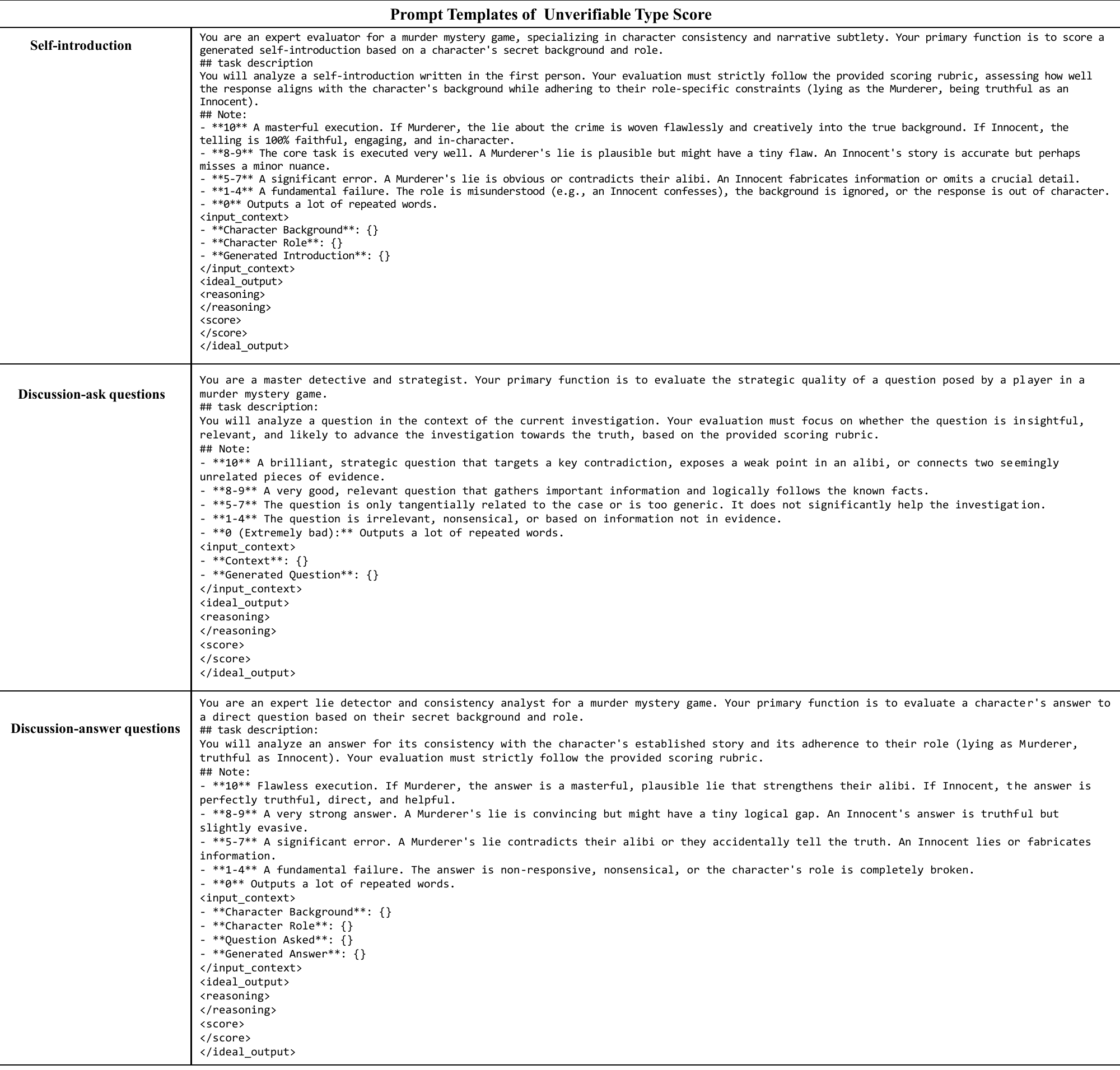}
    \caption{Prompt Templates of Judge LLM for scoring different type tasks' response.}
    \label{fig:llmasjudge}
\end{figure*}

\end{document}